\documentclass[10pt,twocolumn,letterpaper]{article}

\usepackage{iccv}
\usepackage{times}
\usepackage{epsfig}
\usepackage{graphicx}
\usepackage{amsmath}
\usepackage{amssymb}
\usepackage{multirow}
\usepackage{changepage}

\usepackage{booktabs}
\usepackage{makecell}
\usepackage{tabu}
\usepackage{caption}
\usepackage[figuresright]{rotating}
\usepackage{afterpage}
\usepackage{float}
\usepackage{subcaption}

\usepackage[pagebackref=true,breaklinks=true,letterpaper=true,colorlinks,bookmarks=false]{hyperref}

\iccvfinalcopy 

\def\iccvPaperID{****} 

\ificcvfinal\fi

\begin{document}

\title{A Large-scale Study of Spatiotemporal Representation Learning with a New Benchmark on Action Recognition}

\author{Andong Deng$^*$ \qquad Taojiannan Yang\thanks{Equal Contribution.} \qquad Chen Chen\\
Center for Research in Computer Vision\\
University of Central Florida, USA\\
{\tt\small \{dengandong, taoyang1122\}@knights.ucf.edu, chen.chen@crcv.ucf.edu}
}

\maketitle
\ificcvfinal\thispagestyle{empty}\fi

\begin{abstract}
The goal of building a benchmark (suite of datasets) is to provide a unified protocol for fair evaluation and thus facilitate the evolution of a specific area. 
Nonetheless, we point out that existing protocols of action recognition could yield partial evaluations due to several limitations. To comprehensively probe the effectiveness of spatiotemporal representation learning, we introduce \includegraphics[scale=0.07]{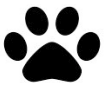}BEAR, a new \textbf{BE}nchmark on video \textbf{A}ction \textbf{R}ecognition. BEAR is a collection of 18 video datasets grouped into 5 categories (anomaly, gesture, daily, sports, and instructional), which covers a diverse set of real-world applications. 
With BEAR, we thoroughly evaluate 6 common spatiotemporal models pre-trained by both supervised and self-supervised learning. We also report transfer performance via standard finetuning, few-shot finetuning, and unsupervised domain adaptation. Our observation suggests that the current state-of-the-art cannot solidly guarantee high performance on datasets close to real-world applications, and we hope BEAR can serve as a fair and challenging evaluation benchmark to gain insights on building next-generation spatiotemporal learners. \textcolor{magenta}{Our dataset, code, and models are released at: https://github.com/AndongDeng/BEAR}
\end{abstract}
\section{Introduction}
\label{sec:intro2}

Learning good spatiotemporal representations~\cite{qian2021spatiotemporal,wu2021mvfnet,han2020self,tong2022videomae, fbvideomae, yang2023aim} is fundamental for video understanding tasks. In action recognition, a common evaluation protocol is to first evaluate the model performance on large-scale video datasets such as Kinetics-400 \cite{kay2017kinetics}, then show its effectiveness of transfer learning to different downstream tasks~\cite{carreira2017quo,feichtenhofer2019slowfast,lin2019tsm,bertasius2021space,liu2022video,text4vis}. 
Many video datasets~\cite{soomro2012ucf101,hmdb51,kay2017kinetics,goyal2017something, epickitchen} have been introduced over the past few years to advance the field. However, there are several major limitations: (1) These datasets are similar in terms of domains and actions. Most of them only contain daily or sports actions because these categories are easy to collect from the web. Yet many important real-world applications, such as anomaly detection and industrial inspection, are rarely included. (2) Each of these datasets has its own characteristics (\eg appearance-focused \cite{kay2017kinetics}, motion-focused \cite{goyal2017something}, fine-grained \cite{shao2020finegym}, egocentric \cite{epickitchen}). Previous works usually conduct evaluations on a few datasets. However, without evaluating a suite of datasets, we cannot fully diagnose a model and make further improvements. (3) The held-out test set for these datasets either does not exist or is not commonly adopted. This will affect the transfer performance because models tuned on a test set using hyperparameter optimization or neural architecture search might achieve good performance but cannot transfer well due to overfitting.

\begin{figure}[t]
\vspace{-5pt}
\centering
    \includegraphics[width=\linewidth]{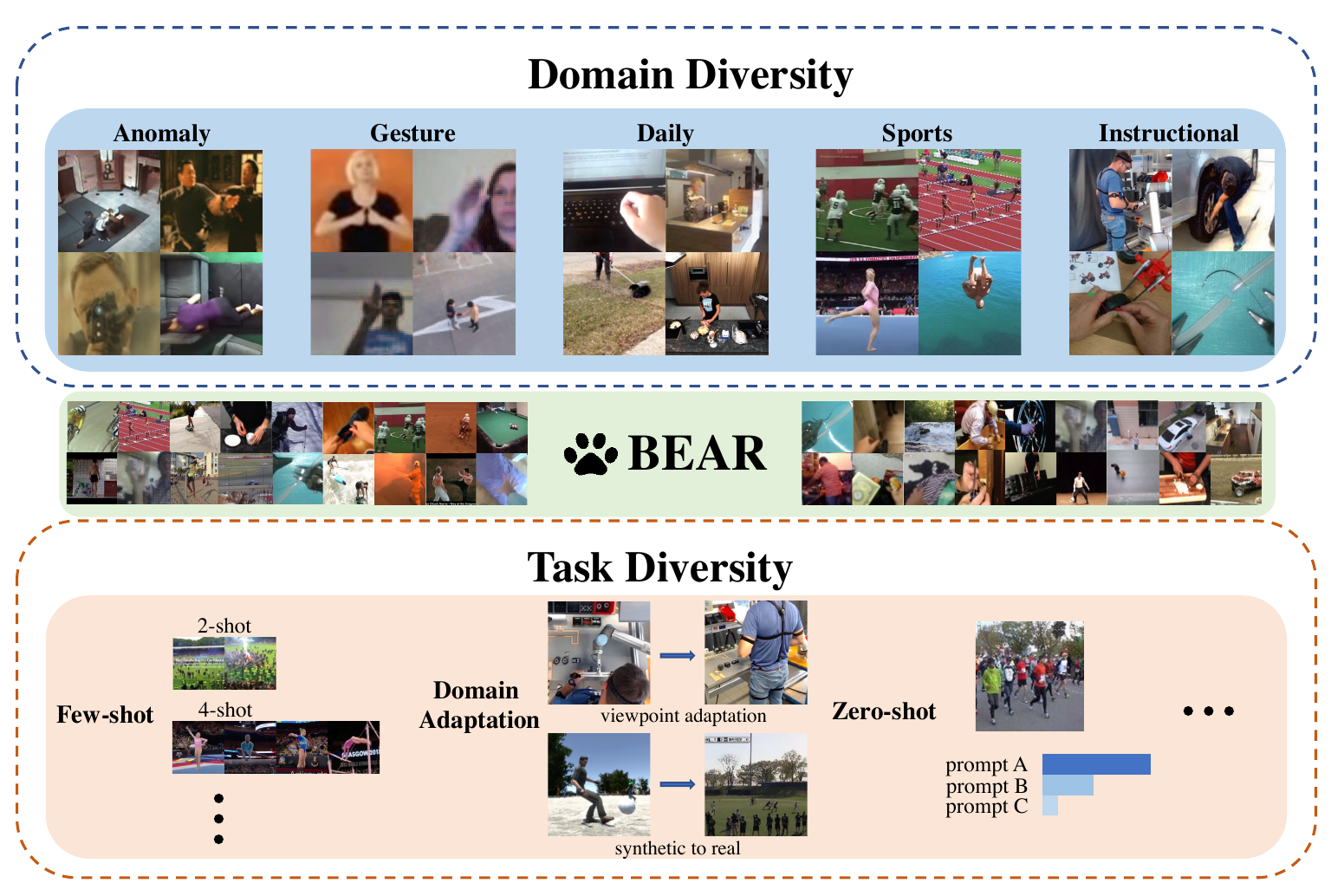}
    \caption{\includegraphics[scale=0.07]{figures/bear.png}BEAR is a collection of 18 video action recognition datasets grouped into 5 categories (Anomaly, Gesture, Daily, Sports, and Instructional). It enables various evaluation settings, e.g., standard finetuning, few-shot finetuning, unsupervised domain adaptation, and zero-shot learning.}
    \label{fig:teaser}
    \vspace{-1.5 em}
\end{figure}

In light of this, we propose a unified and challenging BEnchmark on video Action Recognition, named BEAR, to better evaluate spatiotemporal representation learning. We define good representations as those that can achieve strong transfer learning performance on diverse, unseen domains even with limited data. To this end, we build BEAR by collecting a suite of 18 video action recognition datasets grouped into 5 categories (Anomaly~\cite{sultani2018real,10185076}, Gesture~\cite{materzynska2019jester}, Daily~\cite{sigurdsson2018charades,das2019toyota}, Sports~\cite{karpathy2014large}, and Instructional~\cite{tang2019coin}), which cover a diverse set of real applications. The datasets in BEAR are also diverse in video sources (\eg YouTube, CCTV cameras, self-collected) and viewpoints (\eg egocentric, 3rd person, drone, and surveillance). In addition, we split each dataset into train and test sets, \textit{strictly keeping the test set held out during training in all of our experiments}. 
We will also provide an online evaluation server to enable fair comparisons. 

With BEAR, one can probe spatiotemporal representation learning methods from a much more diverse perspective and answer many important questions. Does the good performance on commonly-used large-scale datasets translate to real applications? Do recent transformer-based models consistently outperform simple 2D models in different domains? How sensitive is the model to domain and viewpoint change? Could the model achieve good performance when downstream data is limited? In this work, we comprehensively investigate 6 representative video models pre-trained by both supervised and self-supervised learning in various settings (\eg full-shot, few-shot, domain adaptation). Our study quantifies existing intuition and uncovers several new insights: 
(1) Simple 2D video models can outperform recent transformer-based models when equipped with strong backbones.
(2) The previous evaluation protocols are constrained to downstream datasets that resemble Kinetics-400. However, the high performance of these datasets does not necessarily transfer to other application domains.
(3) Viewpoint shift has a dramatic impact on downstream task performance. Even the recent domain adaptation methods cannot address the problem to satisfactory. This suggests we may need to go beyond domain adaptation and shift attention to building more comprehensive pre-training datasets.
(4) Self-supervised spatiotemporal representation learning still lags remarkably behind supervised learning. Even the SoTA VideoMAE \cite{tong2022videomae} fails to outperform simple supervised models in diverse domains. 
Our goal is to provide a unified and challenging evaluation benchmark to evaluate spatiotemporal representation learning from various perspectives, which hopefully could guide future development in video understanding. 
\begin{table*}[t]
\begin{center}
\caption{Statistics of the selected datasets used in our video benchmark. We collect 18 datasets covering 5 common data domains for comprehensive benchmarking. In the column of video viewpoint, ``sur.'' means surveillance videos, and ``dro.'' means drone videos. }
\scalebox{0.75}{
\label{table:data_info}
\begin{tabular}{c|ccccccccc}
\hline
\textbf{Dataset}   & \textbf{Domain}  & \textbf{\begin{tabular}[c]{@{}c@{}}Label\\ classes\end{tabular}} & \textbf{\begin{tabular}[c]{@{}c@{}}Clip\\ \textit{num.}\end{tabular}}  & \textbf{\begin{tabular}[c]{@{}c@{}}Avg Length\\ (sec.)\end{tabular}} & \textbf{\begin{tabular}[c]{@{}c@{}}Training data\\ per class (min, max)\end{tabular}} &  \textbf{\begin{tabular}[c]{@{}c@{}}Split\\ ratio\end{tabular}} & \textbf{\begin{tabular}[c]{@{}c@{}}Video\\ source\end{tabular}} & \textbf{\begin{tabular}[c]{@{}c@{}}Video\\ viewpoint\end{tabular}}\\ \hline
XD-Violence~\cite{wu2020not}           & Anomaly       &   5   & 4135     &   14.94    & (36, 2046)    & 3.64:1         & Movies, sports, CCTV, etc.         & 3rd, sur. \\
UCF Crime~\cite{sultani2018real}             & Anomaly       &   12  & 600       &   132.51   & 38    & 3.17:1                 & CCTV Camera                        & 3rd, sur. \\
MUVIM~\cite{denkovski2022multi}                 & Anomaly       &   2   & 1127      &   68.1     & (296, 604)    & 3.96:1        & Self-collected                     & 3rd, sur. \\ \hline
WLASL100~\cite{li2020word}              & Gesture       &   100 & 1375     &   1.23        & (7, 20)    & 5.37:1         & Sign language website              & 3rd \\
Jester~\cite{materzynska2019jester}                & Gesture       &   27  & 133349  &   3        & (3216, 9592)  & 8.02:1          & Self-collected                     & 3rd \\
UAV Human~\cite{li2021uav}             & Gesture       &  155  & 22476   &   5        & (20, 114)   & 2:1               & Self-collected                     & 3rd, dro. \\ \hline
CharadesEgo~\cite{sigurdsson2018charades}      & Daily         &  157  & 42107  &  10.93    & (26, 1120)   & 3.61:1             & YouTube                            & 1st \\
Toyota Smarthome~\cite{das2019toyota}      & Daily         &   31  & 14262      &   1.78    & (23, 2312)    & 1.63:1        & Self-collected                     & 3rd, sur. \\
Mini-HACS~\cite{zhao2019hacs}             & Daily         &  200  & 10000   & 2        & 50       & 4:1                    & YouTube                            & 1st, 3rd\\
MPII Cooking~\cite{rohrbach2012database}          & Daily         &   67  & 3748          & 153.04    & (5, 217)   & 4.69:1        & Self-collected                     & 3rd \\ \hline
Mini-Sports1M~\cite{sports1m}         & Sports        &  487  & 24350     & 10 & 50       & 4:1                        & YouTube                            & 3rd \\
FineGym99~\cite{shao2020finegym}             & Sports        &  99   & 20389     & 1.65        & (33, 951) & 2.24:1            & Competition videos                 & 3rd \\
MOD20~\cite{perera2020multiviewpoint}                 & Sports        &   20  & 2324     & 7.4        & (73, 107)   & 2.29:1           & YouTube and self-collected         & 3rd, dro. \\ \hline
COIN~\cite{tang2019coin}                  & Instructional &  180  & 10426  & 37.01        & (10, 63)    & 3.22:1            & YouTube                            & 1st, 3rd\\
MECCANO~\cite{ragusa2021meccano}               & Instructional &   61  & 7880         & 2.82       & (2, 1157)    & 1.79:1      & Self-collected                     & 1st \\
INHARD~\cite{dallel2020inhard}                & Instructional &   14  & 5303        & 1.36  & (27, 955)    & 2.16:1            & Self-collected                     & 3rd \\
PETRAW~\cite{huaulme2022peg}                & Instructional &    7  & 9727      & 2.16       & (122, 1262)       & 1.5:1    & Self-collected                     & 1st \\
MISAW~\cite{huaulme2021micro}                 & Instructional &   20  & 1551     & 3.8       & (1, 316)   & 2.38:1             & Self-collected                     & 1st \\ \hline
\end{tabular}
}
\end{center}
\vspace{-1.5em}
\end{table*}

\section{Related Work}
\label{sec:related}

\noindent \textbf{Human action recognition} is to distinguish the ongoing actions (or sometimes events) in a video. Different from image classification, video action recognition requires effective temporal modeling~\cite{wang2016temporal}, awareness of the action hierarchies~\cite{shao2020finegym}, and the interaction between the subjects and objects~\cite{goyal2017something}. 
In early years, video models simply inherit the 2D convolution structures~\cite{vgg,resnet} and process temporal information either by extending 2D convolutions into 3D~\cite{tran2015learning,carreira2017quo,yang2021mutualnet} or including optical flow~\cite{simonyan2014two}. However, optical flow-based approaches suffer from costly flow pre-computation, thus 2D CNNs with more sophisticated temporal modeling are designed~\cite{wang2016temporal,zhu2018hidden,lin2019tsm, tsqnet, wu2019multi}.
For 3D CNNs, factorized architectures~\cite{p3d,tran2018closer,s3d,eco} are introduced to improve the model efficiency and reduce overfitting. 
Recently, Transformer~\cite{vaswani2017attention} continues to showcase its capability from language to image and also to video~\cite{bertasius2021space,arnab2021vivit,zhang2021vidtr,liu2022video,yu2023self,wu2021star}. Top performers on most video action recognition datasets are transformer-based. In this work, we fairly evaluate 6 popular video models belonging to 2D CNN, 3D CNN, and Transformer, respectively. With comparable backbones, we surprisingly reveal that 2D CNNs can sometimes outperform transformer models.

\noindent \textbf{Spatiotemporal representation learning} is advancing rapidly in the last few years, especially in a self-supervised manner.
Self-supervised pre-training is appealing because it could learn visual knowledge from massive unlabeled data, which alleviates the annotation burden compared with its supervised counterpart. 
Most approaches design a pretext task to learn the intrinsic spatiotemporal feature within the video data, such as sorting the shuffled video sequence~\cite{lee2017unsupervised}, next frame prediction~\cite{han2019video}, predicting the frame rate~\cite{epstein2020oops}, contrastive learning~\cite{ding2021motion,rhomoco,videomoco,vclr,qian2021spatiotemporal}, mask modeling~\cite{tong2022videomae,fbvideomae}, etc. 
Despite their promising performance, a recent work~\cite{thoker2022severe} points out that video self-supervised pre-training is less robust than its supervised counterpart when the downstream setting varies.
In this work, we also compare supervised pre-training with self-supervised ones in terms of both standard finetuning and few-shot finetuning on our benchmark. 

\noindent \textbf{Vision benchmark} is often designed as a testbed, which consists of multiple datasets from different domains. 
Each benchmark might have its own motivation, but they share the same goal of providing a unified protocol for evaluation and thus facilitating the evolution of a specific area. 
Many well-established benchmarks have been proposed in different research areas ~\cite{wang2019towards,zhai2019visual,li2022elevater,li2021grounded,gupta2022grit}. However, there is no such comprehensive benchmark for video action recognition. 
Two works that are the closest to ours are VTAB~\cite{zhai2019visual} and SEVERE-benchmark~\cite{thoker2022severe}.
VTAB contains 19 datasets that cover a broad spectrum of domains and semantics. 
All tasks are formulated as the image classification problem for the sake of a homogeneous task interface.
Inspired by VTAB, we build the first comprehensive evaluation benchmark for video action recognition.
BEAR includes 18 datasets across 5 domains towards real applications.
It enables fair comparison and thorough investigation of existing video models, which allows us to address interesting open questions.
SEVERE-benchmark~\cite{thoker2022severe} investigates how sensitive video self-supervised learning is to the current conventional benchmark in terms of domain, samples, actions, and tasks. 
Compared to SEVERE-benchmark~\cite{thoker2022severe}, we study both supervised and self-supervised learning in more domains (anomaly, instructional), with more datasets (18 vs 8) and more settings (few-shot, zero-shot, and unsupervised domain adaptation).
\section{\includegraphics[scale=0.1]{figures/bear.png}BEAR}
\label{sec:datasets}

Despite new datasets being introduced every year, the most widely adopted benchmarks in the video action recognition community are Kinetics-400/600/700~\cite{k600,k700,kay2017kinetics}, Something-something-v1/v2~\cite{goyal2017something}, UCF-101~\cite{soomro2012ucf101} and HMDB-51~\cite{hmdb51}. 
However, these datasets share a high similarity in that they are mostly composed of daily and sports actions. 
Models that achieve good performance on these datasets may not generalize well to the challenging real-world scenarios due to dramatic domain shifts. 
For example, anomaly videos are often captured from surveillance cameras, which look quite different from daily videos due to viewpoint change. 
Ideally, a video model is expected to cope with diverse real-world applications.

To comprehensively evaluate the generalization capability of video models, we present BEAR, a new benchmark for human action recognition.
As shown in Table~\ref{table:data_info}, BEAR is a collection of 18 action recognition datasets, carefully designed towards \textit{practical use}, \textit{data diversity}, and \textit{task diversity}. Compared to existing video action recognition datasets, BEAR has the following desirable properties.

\noindent \textbf{Real Applications.} Besides the common daily and sports categories, BEAR contains another three categories including anomaly activity, gesture, and instructional actions. These action categories have important real-world applications such as people fall detection (\eg MUVIM~\cite{denkovski2022multi}), sign language recognition (\eg WLASL100~\cite{li2020word}), industrial inspection (\eg MECCANO~\cite{ragusa2021meccano}), and surgical workflow recognition (\eg  PETRAW~\cite{huaulme2022peg}).

\noindent  \textbf{Data Diversity.} BEAR is not only diverse in application domains but also in the data source, video viewpoint, and video length. As shown in Table~\ref{table:data_info}, BEAR contains videos from various sources such as movies, CCTV cameras, YouTube, and drone cameras. It also includes videos in the 1st and 3rd person views. In terms of video length, the average clip duration varies from the shortest (\eg 1.23s in WLASL100~\cite{li2020word}) to the longest (\eg 153.04s in MPII Cooking~\cite{rohrbach2012database}).
In addition, the training sample size per class varies across datasets, from the lowest (\eg 1 for MISAW~\cite{huaulme2021micro}) to the highest (\eg 9592 for Jester~\cite{materzynska2019jester}).

\noindent \textbf{Few-shot Transfer.} The standard finetuning protocol for transfer is to train a model on the whole training data, which is often more than thousands of videos. 
However, in many real applications, the annotated video data is scarce, \eg anomaly recognition (rarely happens and is costly to label), medical operation (privacy concern), and industrial operation (need the expertise to label). 
To better evaluate a model's potential in real applications, we need to evaluate its effectiveness under few-shot learning.
Hence in BEAR, besides the full datasets, we also split each dataset into 16-shot, 8-shot, 4-shot, and 2-shot versions. This allows researchers and practitioners to thoroughly evaluate a model's sensitivity to data scarcity.

\noindent \textbf{Flexible Evaluation.} Thanks to the data diversity in BEAR, researchers can easily evaluate video models under various settings. For example, full-shot and few-shot learning, domain adaptation from one dataset (or category) to another. Moreover, we also believe that new settings can be easily derived based on our benchmark.

\noindent \textbf{Fair Comparison.} 
The held-out test set for most video action recognition datasets either does not exist or is not commonly adopted. This allows previous methods to conduct hyperparameter optimization or even neural architecture search directly on the test set. Test set tuning usually leads to good testing performance, but it may not translate to other datasets. \textit{To promote fair comparison and generalization capability, we will hold the test sets and provide an evaluation server for future researchers and practitioners.}

\noindent \textbf{Dataset Accessibility.} We provide scripts to download and format all 18 datasets automatically. Our codebase is built upon MMAction2~\cite{2020mmaction2}, so researchers can easily integrate their new models by providing a model definition file without additional efforts to perform evaluations. Furthermore, the total number of video clips in BEAR is about 310K, which is comparable to Kinetics-400. \textit{Therefore, the overall time cost is similar to training a model on Kinetics-400.}

\section{Models}
There has been a considerable amount of video models proposed to solve the human action recognition task. 
From the perspective of the basic building block, these models can be roughly classified into three categories: 2D CNNs, 3D CNNs, and transformer-based models. 
To investigate the efficacy of each model type, in this work, we select two representative works from each category: TSN~\cite{wang2016temporal} and TSM~\cite{lin2019tsm} for 2D CNNs, I3D~\cite{carreira2017quo} and 3D Non-local network~\cite{wang2018non} for 3D CNNs, TimeSformer~\cite{bertasius2021space} and VideoSwin~\cite{liu2022video} for transformer-based models. 
We would like to point out that for CNN-based models (TSN, TSM, I3D, and NL), we choose ConvNext-base~\cite{liu2022convnet} as the backbone because it has a similar model size and performance, as shown in Table~\ref{backbone},  on ImageNet-1K compared to ViT-B and Swin-B, which is the backbone of TimeSformer and VideoSwin, respectively. 
This alleviates the impact from the backbone, thus presenting a more fair comparison among different video architectures. 
In this work, we finetune all the models based on both supervised and self-supervised pre-training on Kinetics-400, and the pre-training performance is shown in Table \ref{tab:pretrained}. The pre-training details can be found in \textcolor{red}{Appendix~\ref{appendix training details}}.
In the following sections, we will provide a comprehensive study w.r.t. transferring performance from multiple perspectives: standard finetuning, few-shot finetuning, unsupervised domain adaptation, and zero-shot evaluation.

\begin{table}[h]
\centering
\caption{Comparison among ConvNeXt-base, ViT-base, and Swin-base. Params denote the parameters volume, and Top-1 acc means the top-1 accuracy in the ImageNet classification task.}
\scalebox{1.}{
\setlength{\tabcolsep}{1mm}{
\begin{tabular}{c|cc}
\hline
Backbone               & Params(M)        & Top-1 acc($\%$) \\\hline
ConvNeXt-base~\cite{liu2022convnet}        & 88.59               & 85.8          \\ 
ViT-base~\cite{dosovitskiy2020image}              & 86.57              & 81.8       \\
Swin-base~\cite{liu2021swin}             & 87.77              & 85.2          \\ \hline
\end{tabular}}}

\label{backbone}
\end{table}

\begin{table}[t]
\caption{The pre-training results of 6 models on Kinetics-400 in both supervised and self-supervised settings. The supervised results are based on the single-view test, and the self-supervised ones are based on KNN evaluation.}
\centering
\small
\setlength{\tabcolsep}{6mm}{
\begin{tabular}{c|cc}
\hline
model           & Supervised       & SSL \\\hline
TSN             & 77.6               & 43.1  \\
TSM             & 76.4               & 43.2  \\
I3D             & 74.2                & 51.3    \\
NL     & 73.9               & 50.7  \\
TimeSformer     & 75.8               & 50.3  \\
VideoSwin      & 77.6               & 51.1  \\ \hline
\end{tabular}}

\vspace{-1.5em}
\label{tab:pretrained}
\end{table}

\section{Standard Finetuning}
\label{sec:finetune}

Finetuning models that are pre-trained on large-scale datasets have been a mainstream learning paradigm in deep learning, and performance on various downstream datasets can provide a more comprehensive evaluation with less bias. Thus, in BEAR, we regard standard finetuning as a basic evaluation method. Specifically, we finetune the pre-trained models on the 18 datasets to investigate: 
1) the performance of different types of video models on different data domains; 2) the difference between supervised pre-training and self-supervised pre-training; 3) potential factors (\eg domain shift, viewpoint shift, etc.) that have significant impacts on the performance of downstream tasks. 
We want to emphasize that during finetuning, we do not tune hyperparameters on the test set to avoid potential overfitting. All reported results are based on the evaluation of the last checkpoint. The Top-1 accuracy of each model is presented in Table~\ref{tab:finetune}. Besides the performance on each dataset, we also propose two \textit{composite metrics} over the 18 datasets for evaluation. The first one is the macro-average accuracy which is the average of the accuracy on each dataset. The second one is micro-average accuracy, which calculates the average accuracy on the video level. Micro-average considers the size difference of the 18 datasets. We include the details of the complete finetuning results, and the previous best-reported performance, if any, for each dataset in \textcolor{red}{Appendix~\ref{appendix full}}.

\begin{table*}[!t]
    \caption{Finetuning results based on the supervised pre-trained and self-supervised pre-trained models as well as the X3D pre-trained models. Generally, from the two composite metrics (macro-average accuracy and micro-average accuracy), we can tell that TSM surprisingly outperforms other counterparts in both pre-training settings.}\label{tab:finetune}
    \centering
    \scalebox{0.72}{
    \begin{tabular}{c|cccccc|c|cccccc}
          \hline
          \multirow{2}*{\textbf{Dataset}} & 
          \multicolumn{6}{c|}{\textbf{Supervised pre-training}} & 
          \multirow{2}*{\textbf{X3D}} &
          \multicolumn{6}{c}{\textbf{Self-supervised pre-training}}  \\
          \cline{2-7}
          \cline{9-14}
                                      &  \textbf{TSN} &\textbf{TSM} & \textbf{I3D} & \textbf{NL} & \textbf{TimeSformer} & \textbf{VideoSwin}   & ~      & \textbf{TSN} &\textbf{TSM} & \textbf{I3D} & \textbf{NL} & \textbf{TimeSformer} & \textbf{VideoSwin}   \\ \hline
         \textbf{XD-Violence}         & \textbf{85.54}         & 82.96       & 79.93       & 79.91       & 82.51                & 82.40       & 75.11        & 80.49        & \textbf{81.73}       & 80.38        & 80.94       & 77.47                & 77.91                \\ 
         \textbf{UCF-Crime}           & 35.42         & \textbf{42.36}       & 31.94         &  34.03      & 36.11                & 34.72     & 25.69          & \textbf{37.50}        & 35.42       & 34.03        &  34.72      &   36.11             & 34.03               \\ 
         \textbf{MUVIM}               & 79.30         & \textbf{100}         &  97.80      &  98.68      & 94.71        &  \textbf{100}     & 99.56          & 99.12        & \textbf{100}         & 66.96        & 66.96       & 99.12                & \textbf{100}  \\ \hline
         \textbf{WLASL}               & 29.63        & 43.98       &   49.07      &  \textbf{52.31}    & 37.96                & 45.37         & 44.91       & 27.01        & 27.78       & 29.17        & \textbf{30.56}       &   25.56             & 28.24            \\ 
         \textbf{Jester}              & 86.31         & \textbf{95.21}       & 92.99        &  93.49      & 93.42                & 94.27      & 92.24         & 83.22        & \textbf{95.32}       & 87.23        &  93.89      & 90.33                & 90.18              \\ 
         \textbf{UAV-Human}           & 27.89         & \textbf{38.84}       &  33.49       & 33.03         & 28.93                & 38.66    & 36.07          & 15.70        & 30.75       & 31.95        &26.28        & 21.02               & \textbf{35.12}          \\ \hline
         \textbf{CharadesEGO}         & 8.26          & 8.11        &  6.13        & 6.42         & \textbf{8.58}                 & 8.55      & 5.69   & 6.29         & 6.59        & 6.24         & 6.31        & 7.59            & \textbf{7.65}             \\ 
         \textbf{Toyota Smarthome}    & 74.73         & \textbf{82.22}       & 79.51        & 76.86      & 69.21                & 79.88       & 79.09          & 68.71        & \textbf{81.34}       & 77.82        & 76.16       & 61.64                  & 80.18              \\ 
         \textbf{Mini-HACS}           & 84.69         & 80.87       &77.74         &  79.51      & 79.81                & \textbf{84.94}       & 60.57          & 64.60        & 63.24       & 70.24        & 60.57       & 73.92                  &  \textbf{75.58}           \\ 
         \textbf{MPII Cooking}        & 38.39         & 46.74       & \textbf{48.71}         &    42.19    & 40.97                & 46.59      &  42.19        & 34.45        & \textbf{50.08}       & 42.79        & 40.36       & 35.81               & 47.19            \\ \hline
         \textbf{Mini-Sports1M}       & 54.11         & 50.06       &46.90         & 46.16          & 51.79                & \textbf{55.34}     & 41.91         & 43.02        & 43.59       & 46.28        & 45.56       & 44.60                & \textbf{47.60}              \\ 
         \textbf{FineGym}             & 63.73         & \textbf{80.95}       &   72.00         & 71.21         & 63.92                & 65.02   & 68.49           & 54.62        & \textbf{75.87}       & 69.62        & 68.79       & 47.60                & 58.94              \\ 
         \textbf{MOD20}               & \textbf{98.30}         & 96.75       &  96.61      &  96.18        & 94.06                & 92.64       &  92.08        & 91.23        & 92.08       & 91.94        & 92.08       & 90.81               & \textbf{92.36}             \\ \hline
         \textbf{COIN}                & 81.15         & 78.49       & 73.79        & 74.30        & \textbf{82.99}                & 76.27       & 61.29       & 61.48        & 64.53       & 71.57        & \textbf{72.78}       & 67.64                  & 68.78              \\ 
         \textbf{MECCANO}             & \textbf{41.06}         & 39.28       & 36.88        & 36.13          & 40.95                & 38.89      & 30.78         & 32.34         & 35.10        & 34.86        & 33.62        & 33.30              & \textbf{37.80}              \\ 
         \textbf{InHARD}              & 84.39         & \textbf{88.08}       &  82.06      & 86.31       & 85.16                & 87.60          & 84.86      & 75.63          & \textbf{87.66}         & 82.54        & 80.81         & 71.28           & 80.10              \\ 
         \textbf{PETRAW}              & 94.30         & 95.72       & 94.84        & 94.54      & 94.30                & \textbf{96.43}          & 95.46    & 93.18        & \textbf{95.51}       & 95.02        & 94.38       & 85.56                  &  91.46             \\ 
         \textbf{MISAW}               & 61.44         & \textbf{75.16}       &  68.19     & 64.27        & 71.46                & 69.06         & 69.06     & 59.04        & \textbf{73.64}       & 70.37        & 64.27       & 60.78                  &  68.85             \\ \hline
         Macro Avg.                 & 62.70     & \textbf{68.10}        & 64.92       & 64.75        & 64.27       & 66.48               &  61.39      & 57.09         & \textbf{63.35}       &  60.50     & 59.39        & 57.23                & 62.33               \\ 
         Micro Avg.              & 64.92        & \textbf{70.82}       & 67.81        & 67.83       & 67.66                  &  69.73         & 65.87      & 59.13         & \textbf{68.11}       &  64.35     & 66.21        & 62.19               & 65.71         \\ \hline
    \end{tabular}}
\end{table*}

\paragraph{Model comparison.} 
In previous studies, transformer-based video models~\cite{bertasius2021space,liu2022video} have been demonstrated to be more effective than CNNs on several representative datasets. This conclusion leads the trend of model design toward more sophisticated transformers, which makes CNNs less appealing compared with the pre-transformer era. However, we argue that the current conclusion could be biased since the comparison between transformers and CNNs is obviously unfair. Basically, it is a widely accepted notion that the selection of different backbones can inherently yield significant differences, let alone the overall model design. To this end, as aforementioned, we carefully select ConvNeXt~\cite{liu2022convnet} as the CNN backbone, which is comparable with ViT~\cite{dosovitskiy2020image} and Swin Transformer~\cite{liu2021swin} w.r.t. both model size and ImageNet classification performance. We believe such a fair comparison could lead to more convincing and compelling conclusions. As shown in Table~\ref{tab:finetune}, we notice that there is no absolute winner among all the models, but surprisingly, 2D CNNs perform better on most datasets, especially TSM, which outperforms other models in 8 out of 18 datasets. This indicates that 2D video models are still competitive with transformers when equipped with strong backbones.
Likewise,  the two composite metrics also provide evidence that TSM outperforms other models, and transformer-based models do not exhibit clear advantages over CNN-based models. 

Inspecting further, we can see that VideoSwin excels in mini-HACS and mini-Sports1M. However, as aforementioned, these datasets, along with other popular datasets such as UCF-101 and HMDB-51, share high similarities with Kinetics-400 in terms of actions and viewpoints. Thus the performance on these datasets may not fully reflect the effectiveness of the evaluated model. Indeed, as shown in Table \ref{tab:finetune}, VideoSwin is only comparable or inferior to TSM in the other three categories (\ie, anomaly, gesture, and instructional). This demonstrates that the impressive performance on Kinetics-400 and other similar datasets may not be consistent with downstream tasks with vastly different actions. To fully probe the effectiveness of a video model, we need to evaluate it on datasets with different distributions. Besides, we also consider the NAS-based X3D~\cite{feichtenhofer2020x3d}, which achieves good performance on Kinetics-400, to reveal the overfitting problem of tuning on the test set.

\begin{adjustwidth}{-1.5em}{}
\begin{itemize}
\item \emph{Despite the emergence of recent transformers, 2D video models can still be promising alternatives for action recognition if equipped with powerful backbones.}
\item \emph{Previous evaluation protocols have been limited to target datasets similar to Kinetics-400, which could potentially result in biased evaluations. However, BEAR could address this issue by including target data from five distinct domains, ensuring a more comprehensive and unbiased assessment of model performance.}
\end{itemize}
\end{adjustwidth}

\paragraph{Impact of viewpoint change} 
We also observe something interesting in terms of the data distribution. Several datasets such as UCF-Crime, UAV-Human, CharadesEGO, MPII-Cooking, and MECCANO exhibit notably low performance. Upon closer inspection of Table~\ref{table:data_info}, it is evident that these datasets involve significant viewpoint changes from Kinetics-400. For instance, UCF-Crime is collected from CCTV footage, UAV-Human contains drone-view videos, CharadesEGO only contains 1st person-view videos, and MECCANO is also egocentric. This indicates that the viewpoint change in downstream tasks could dramatically damage the model performance.
Therefore, leveraging pre-training datasets with rich egocentric visual knowledge, such as EGO4D~\cite{grauman2022ego4d}, may offer a suitable alternative to Kinetics-400 for finetuning on egocentric data.
Besides, in Sec.~\ref{sec:few-shot} and Sec.~\ref{sec:uda}, we will further discuss the challenge caused by the viewpoint change in the target domain. 

\begin{adjustwidth}{-1.5em}{}
\begin{itemize}
    \item \emph{Prior evaluation protocols, limited in the scope of target data, fail to capture the impact of domain gap, particularly in regard to the viewpoint, 
    on transfer performance. However, we have identified that such a distribution shift can significantly degrade the quality of spatiotemporal representation, which further undermines the transfer performance. Hence, we recommend that future studies should include pre-training datasets beyond Kinetics-400 to provide more robust representations to improve transferability.}
\end{itemize}
\end{adjustwidth}

\noindent\textbf{Self-supervised vs. supervised pre-training} 
As can be seen from Table~\ref{tab:finetune}, it is notable that the overall finetuning performance of the self-supervised pre-training is less competitive than its supervised counterpart even for TSM. The most pronounced accuracy drop can be found in WLASL and FineGym. The performance of 3D Nonlocal network on WLASL drops from 52.31$\%$ to 30.56$\%$ and the performance of TimeSformer also decreases more than 15$\%$. 
To reveal the potential reason behind this, we further scrutinize the data distribution gap between the selected 18 target datasets and Kinetics-400. We observe different types of domain shifts, such as UAV-Human containing only drone-view data and the egocentric MECCANO which differs significantly from Kinetics-400. We conclude that self-supervised pre-training is more susceptible to domain shifts between Kinetics-400 and the target datasets than supervised pre-training. In Sec.~\ref{sec:few-shot}, we take a step forward on this topic by investigating few-shot settings, which are more likely to occur in real-world scenarios.

\begin{adjustwidth}{-1.5em}{}
\begin{itemize}
    \item \emph{Self-supervised finetuning generally cannot outperform its supervised counterpart and TSM consistently performs well under the self-supervised setting.}
\end{itemize}
\end{adjustwidth}
\begin{figure*}[htbp]
\centering
    \includegraphics[width=17.5cm]{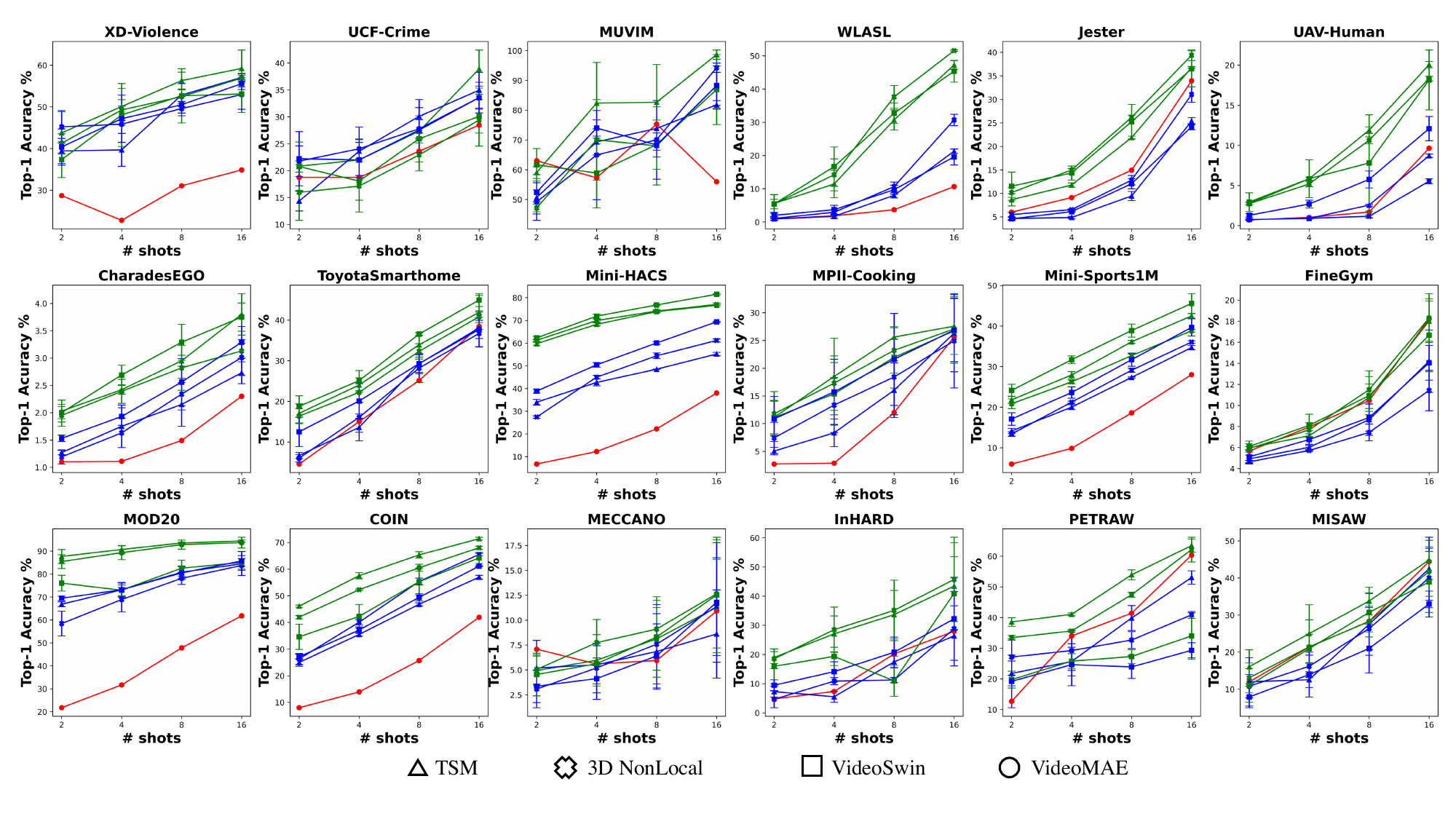}
    \vspace{-2.5em}
    \caption{Results of few-shot learning based on supervised and self-supervised pre-training. The \textcolor[rgb]{0.4,0.8,0.4}{green} curves represent supervised pre-training and the \textcolor[rgb]{0,0,1}{blue} curves represent $\rho$MoCo self-supervised pre-training. We illustrate the results of TSM, 3D NonLocal, and VideoSwin for both pre-training methods. Additionally, we add the SOTA self-supervised pre-training method VideoMAE, represented by the \textcolor[rgb]{1,0,0}{red} curves, for comparison. It could be obvious that even the VideoMAE could lag a lot behind in the few-shot setting.}
    \label{fewshot}
\end{figure*}

\section{Few-shot Learning}
\label{sec:few-shot}

Compared with standard finetuning where abundant annotations can be utilized, few-shot learning is of more practical significance since annotating massive amounts of videos is notoriously expensive. To extend the investigation mentioned in Section \ref{sec:finetune},  we thoroughly investigate the capability of the selected 6 models on BEAR under a few-shot setting given both supervised and self-supervised pre-trained weights. Specifically, we consider (2,4,8,16)-shot settings, and for each setting, we randomly generate 3 splits and report the mean and standard deviation. 
Due to space constraints, we only select TSM, 3D NonLocal, and Video Swin to represent each model type for illustration as they perform generally better.  Complete few-shot results and the training details are in \textcolor{red}{Appendix~\ref{appendix few}}.
\paragraph{Model comparison.} 

The rankings of the six models in few-shot finetuning exhibit distinct variations compared to the standard finetuning. In contrast to the dominance of TSM in standard finetuning across both pre-training settings, the most effective models differ significantly across datasets in few-shot finetuning. Figure~\ref{fewshot} demonstrates that TSM no longer clearly outperforms other models in most datasets, and the two composite metrics (which are presented in the Supplementary due to space limitations) support this conclusion. Specifically, TSM and TimeSformer exhibit similar performance in supervised pre-training, whereas I3D and VideoSwin perform better in self-supervised learning. These findings further reveal the limitations of previous simple evaluation protocols, which may not provide a fair assessment of video models.
These results also confirm the necessity of BEAR, which emphasizes the importance of diverse downstream datasets and various settings for unbiased evaluation. 

\begin{adjustwidth}{-1.5em}{}
\begin{itemize}
    \item \emph{The ranking relations between models could exhibit differently between standard and few-shot finetuning even within the same datasets. This finding further emphasizes the importance of our proposed BEAR benchmark, which advocates for a comprehensive evaluation approach that considers both dataset diversity and finetuning settings.}
\end{itemize}
\end{adjustwidth}

\paragraph{Impact of viewpoint change} 
As in standard finetuning, viewpoint change also has a severe impact when it comes to few-shot learning.
Comparing the results in Figure \ref{fewshot} with those in Table \ref{tab:finetune}, we can see that
the few-shot learning performance decreases drastically in general, especially in datasets that have less in common with Kinetics-400, such as UAV-Human, which is constructed by videos captured from unmanned aerial vehicles, FineGym, which contains fine-grained gym-related videos, and PETRAW and MISAW, which are simulated medical operations in the 1st person view. Conversely, in datasets that are more similar to Kinetics-400, these performance gaps are notably reduced. For example, even the 2-shot performance on Mini-HACS and MOD20 can reach approximately 60\% and 85\%, and the models achieve satisfying performance on the 16-shot setting on COIN.
In previous works, the homogeneity of the pre-training and downstream data hindered the timely identification of such phenomena in few-shot learning. Our investigation highlights the challenge of few-shot learning and underscores the importance of bridging the gap (as aforementioned, introducing extra data, such as Ego4D) between pre-training and the target data. 

Moreover, in the few-shot setting, self-supervised pre-training is more susceptible to viewpoint change. In challenging datasets such as UAV-Human and WLASL, few-shot learning can hardly obtain satisfying results based on self-supervised pre-trained weights, while in the 16-shot setting, supervised pre-training could provide comparable performance compared with standard finetuning. Similarly, in MOD20, the performance experiences a sharp decline in few-shot settings with self-supervised pre-training, while supervised pre-trained TSN and TSM can achieve accuracy exceeding 90\% in the 16-shot.

\begin{adjustwidth}{-1.5em}{}
\begin{itemize}
    \item \emph{
    Few-shot finetuning remains a significant challenge in real-world scenarios.  The performance drops dramatically compared to standard finetuning especially when there is a large domain gap between pre-training and target data. However, when downstream datasets are similar to source data, the performance drop could be mitigated.
    }
    \item \emph{In few-shot learning, self-supervised pre-training is more vulnerable to viewpoint shift, while supervised pre-trained models can achieve favorable performance compared with standard finetuning on the 16-shot setting.}
\end{itemize}
\end{adjustwidth}

\paragraph{Self-supervised vs. supervised pre-training.} 
Comparing the blue curves to the green curves in Figure~\ref{fewshot}, we can see that self-supervised pre-training is generally less effective than supervised pre-training, which is consistent with the conclusion in Sec.~\ref{sec:finetune}. The performance gaps are pronounced in gesture datasets and are less significant in Mini-Sports1M, ToyotaSmarthome, etc. The performance gap is also different across different models. The largest gap appears in TSN and TimeSformer (the complete results are provided in \textcolor{red}{Appendix Table~\ref{tab:sup res1}-\ref{tab:ssl res3}}).
One reason for the poor performance of self-supervised learning may be the limitation of $\rho$MoCo. Therefore,
to consolidate our conclusion,
we further consider VideoMAE~\cite{tong2022videomae}, which is the SoTA self-supervised method and has demonstrated even better performance than supervised models on multiple datasets. Here, we use the officially released VideoMAE ViT-B model, which achieves 81.5\% Top-1 accuracy on Kinetics-400. However, comparing the results with our 6 supervised pre-trained models in Figure \ref{fewshot} (red vs. green curves), we show that VideoMAE could only be comparable with the best supervised pre-trained models in less than half of the datasets.

\begin{adjustwidth}{-1.5em}{}
\begin{itemize}
    \item \emph{Supervised pre-training shows consistent advantages over self-supervised ones in few-shot finetuning. Even the SoTA VideoMAE can hardly outperform simple supervised pre-trained models in diverse domains.}
\end{itemize}
\end{adjustwidth} 
\section{Unsupervised Domain Adaptation}
\label{sec:uda}

\begin{table*}[h]
\caption{The unsupervised domain adaptation accuracy on our UDA datasets: Toyota Smarthome-MPII-Cooking (T: Toyota Smarthome, M: MPII-Cooking), Mini-Sports1M-MOD20 (MS: Mini-Sports1M, MOD: MOD20), UCF-Crime-XD-Violence (U: UCF-Crime, X: XD-Violence), PHAV-Mini-Sports1M (P: PHAV, MS: Mini-Sports1M), Jester, InHARD (I: InHARD, T: Top, L: Left, R: Right). }
\scalebox{0.85}{
\setlength{\tabcolsep}{1mm}{
\begin{tabular}{c|cc|cc|cc|c|c|cccccc}
\hline
Settings & \multicolumn{7}{c|}{\textbf{Inter-dataset}} & \multicolumn{7}{c}{\textbf{Intra-dataset}}\\ \hline
Dataset               & T$\rightarrow$M & M$\rightarrow$T & MS$\rightarrow$MOD & MOD$\rightarrow$MS & U$\rightarrow$X & X$\rightarrow$U & P$\rightarrow$MS & Jester & IT$\rightarrow$IL & IT$\rightarrow$IR & IL$\rightarrow$IR & IL$\rightarrow$IT & IR$\rightarrow$IT & IR$\rightarrow$IL \\\hline
Source only                     & 5.32               & 7.36        & 18.25        & 12.76     & 54.20 & 33.33     & 61.45 & 68.73 & 4.18 & 30.39 & 19.01 & 22.65 & 24.14 & 12.42 \\ \hline
TA$^{3}$N~\cite{chen2019temporal}   & 11.17              & 15.38       & 23.77        & 19.15   & 59.91 & 44.44   & 65.79 & 71.44 & 5.78 & 41.83 & 27.91 & 28.08 & 35.66 & 14.68 \\
CoMix~\cite{sahoo2021contrast}   & 12.63              & 15.32       & 24.48        & 21.56     & 60.17 & 47.22    & 64.83 & 75.86 & 6.32 & 39.79 & 30.45 & 31.74 & 32.94 & 14.83 \\ \hline
Supervised target                  & 70.21              & 65.13        & 34.08        & 35.52   & 75.06 & 63.89   & 94.40 & 97.61 & 26.00 & 83.55 & 83.55 & 85.52 & 85.52 & 26.00 \\ \hline
\end{tabular}}}
\label{tab:uda}
\end{table*}

\begin{figure*}[t]
\centering
    \includegraphics[width=\linewidth]{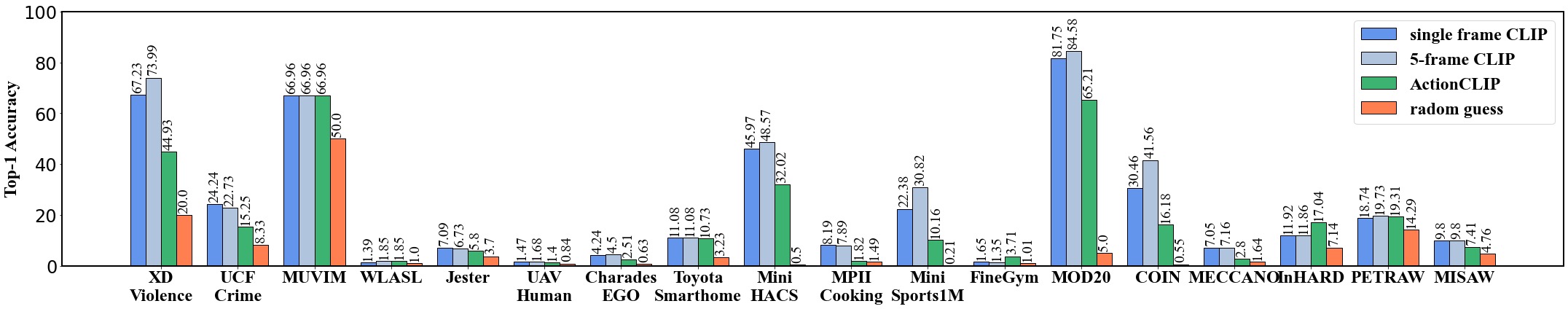}
    \caption{Results of zero-shot evaluation. For most datasets in our benchmark, CLIP-based models still cannot provide reasonable results, especially for those challenging datasets with severe viewpoint shifts and fine-grained datasets.}
    \label{zero}
    \vspace{-2ex}
\end{figure*}

In real-world scenarios, it is possible to transfer knowledge from similar datasets which are well-annotated to others with only limited labels. For instance, there are a lot of existing datasets that include samples of the same categories in the corresponding real-world tasks and thus can be used to facilitate model training. 
Nonetheless,  due to the domain gap, models directly trained on one dataset cannot be well generalized on the target data. In such case, unsupervised domain adaptation (UDA)~\cite{ganin2015unsupervised} can largely alleviate this distribution shift issue by learning the domain-invariant feature when labeled source data is available, learning representations that would promote the performance on the target domain. In BEAR, we construct several dataset pairs for UDA based on two different paradigms: inter-dataset adaptation and intra-dataset adaptation. Given that one of the features of our benchmark is that we collect several datasets with obvious viewpoint shifts, we also focus on this point when we build our UDA datasets. The details of the dataset statistics can be found in \textcolor{red}{Appendix~\ref{appendix uda}}.
We provide two common baseline results: `Source only' and `Supervised target'. The former directly evaluates the model trained on the source training set with the target test set, and the latter is the supervised learning performance on the target domain. Besides, we also evaluate two recent UDA algorithms on our benchmark: TA$^{3}$N~\cite{chen2019temporal} and CoMix~\cite{sahoo2021contrast}. 

\paragraph{Inter-dataset adaptation.}
Inter-dataset is constructed based on two different datasets that have different distributions, especially viewpoint change, but share common categories. Toyota Smarthome contains videos captured from 7 different cameras deployed in an apartment, while MPII-Cooking consists of videos from a down-view camera. Specifically, we select 6 new categories, which contains original action classes in Toyota Smarthome and MPII-Cooking, for the new Toyota Smarthome-MPII-Cooking dataset. The number of  videos is 5,233 and 943 for Toyota Smarthome and MPII-Cooking, respectively. 
Similarly, for Mini-Sports1M and MOD20, we select 15 categories to build the new dataset. In contrast to Toyota Smarthome-MPII-Cooking, the data distribution in Mini-Sports1M-MOD20 is much more balanced. There are 1,650 videos for Mini-Sports1M and 1,767 for MOD20.
We also consider the anomaly detection dataset. Basically, there are three shared action categories in UCF-Crime and XD-Violence: \textit{abuse}, \textit{fighting}, and \textit{shooting}. The domain shift in this dataset is also conspicuous: all the videos in UCF-Crime are from surveillance footage, where the target objects in video frames can only be in a small region, while most videos in XD-Violence are collected from action movies, which could record an action with abundant details. 
To provide a dataset for synthetic-to-real transfer, which is of great significance in real-world scenarios, we also include the simulated dataset PHAV~\cite{roberto2017procedural} to construct PHAV-Mini-Sports1M dataset. We combine 15 classes from Mini-Sports1M into 6 categories (\textit{playing soccer}, \textit{playing golf}, \textit{playing baseball}, \textit{shooting gun}, \textit{shooting archery} and \textit{running}) existing in PHAV to build the paired dataset. 

\paragraph{Intra-dataset adaptation. } 
Intra-dataset, on the contrary, is built within one dataset that records the same actions differently. 
We include Jester(S-T), which is initially introduced by \cite{sahoo2021contrast}, in BEAR since it has been a well-established dataset for domain adaptation. Each identical action in Jester with a contrary direction is merged into one category. 
We also construct a three-view dataset based on InHARD. Basically, each original frame in InHARD contains three distinguished views (i.e., top, left, and right). We simply split the frames according to the view and construct three sub-datasets as InHARD-Top, InHARD-Left, and InHARD-Right. We keep the category the same as the original dataset.

\paragraph{Challenging viewpoint adaptation.}
As shown in Table~\ref{tab:uda}, domain adaptation can be obviously challenging, especially in viewpoint change cases. For instance, ToyotaSmarthome and MPII-Cooking share similar attributes w.r.t. their actions, since they both record kitchen events. However, videos in ToyotaSmarthome are recorded via different cameras in the living room, while videos in MPII-Cooking are recorded by a down-view camera. The performance between these two datasets is far lower than the `supervised target'. Similar observations can also be obtained in InHARD. Although the adaptation is conducted within the dataset, recent methods still fail to perform well when adapting from one viewpoint to another. However, the gap between supervised target and UDA methods is much smaller in other UDA datasets where the viewpoint change is smaller. These results, along with the observations in Secs. \ref{sec:finetune} and \ref{sec:few-shot}, reveal that viewpoint change has a critical impact on transfer performance, which is hard to mitigate even with recent UDA algorithms.

\section{Zero-shot Learning}
\label{sec:zeroshot}
Direct finetuning on annotated datasets is a commonly adopted paradigm for action recognition, but the recent success of vision-language models, which leverage the rich correspondence between natural language and visual content, has provided a new learning paradigm for vision tasks in a zero-shot setting, which is severely required in applications without labeled data. Therefore, we also provide the zero-shot evaluation on BEAR using the recent CLIP-based~\cite{radford2021learning,wang2021actionclip} models.

Basically, we provide two different settings for frame-level CLIP evaluations, \ie single-frame, which follows the settings in \cite{radford2021learning} and 5-frame, where we sample 5 frames from the input video and fuse the model output of each frame. Similarly, we also construct multiple templates for each dataset to obtain ensemble textual embeddings. Considering the inconsistent label domains for the selected datasets, we provide different templates given their distinct attributes of both data and labels. For instance, UCF-Crime~\cite{sultani2018real} is mostly constituted of surveillance videos in a crime scene; thus, a sentence like \textit{`a photo from a surveillance camera showing a criminal doing \{\} in a crime scene.'} is utilized as a part of the prompts. Additionally, we evaluate all the datasets via ActionCLIP \cite{wang2021actionclip}, which is pre-trained on Kinetics-400 based on video and label-text correlation, to unmask the difference of zero-shot performance between image-based models and video-based models. 

As illustrated in Figure \ref{zero}, different from its versatility in the image domain, most of the zero-shot results based on CLIP are still far lower than those of supervised learning. For example, WLASL shows poor correlations between frames and the corresponding labels, which can be partly explained by the large visual gap between the visual information of sign languages and the label itself. Surprisingly, for most datasets, ActionCLIP, which leverages more frames, performs even worse than CLIP. Part of the reason could be that ActionCLIP finetunes CLIP on Kinetics-400, which leads to catastrophic forgetting and overfitting.  However, for some datasets, zero-shot learning could outperform few-shot learning, such as XD-Violence and MOD20, which even approaches supervised learning. This may be partly because the high vision-text correlation existed in these datasets, and this also demonstrates the potential of language supervision in action recognition.

\section{Conclusion and Discussion}
In this work, we introduce a new action recognition benchmark \includegraphics[scale=0.07]{figures/bear.png}BEAR to address several limitations in existing video benchmarks. Aiming at benefiting both academic and industrial applications, we carefully select 18 datasets covering 5 distinct data domains.
Such a wide scope could provide comprehensive assessment protocols for any video model, filling the gap in the current video action recognition benchmark that only a small number of target datasets are considered. It helps prevent models from overfitting on a specific dataset which could result in biased model evaluation. 
Moreover, to achieve a fair comparison, we held out test data for every dataset and avoid using it for parameter selection during training, and the evaluation is based on the last checkpoint. Meanwhile, in this work, we also pay attention to the capabilities of 2D CNNs, 3D CNNs, and transformers. 
Importantly, we carefully select comparative backbones for them to avoid erroneous comparisons. 

Based on our extensive experiments, we have several interesting and instructive observations: 1) 2D video models are competitive with SoTA transformer models when equipped with strong backbones. 2) Previous evaluation protocols on a few similar datasets can yield biased evaluation. 3) Domain shift (especially the viewpoint shift) has a large impact on transfer learning, and the performance gap could be much more remarkable in the few-shot setting. 4) Self-supervised learning still largely falls behind supervised learning, and even the SoTA VideoMAE cannot outperform supervised models on diverse downstream datasets. Moreover, we also point out that in order to learn robust spatiotemporal representations, constructing new pre-training datasets containing videos from diverse domains could benefit the target performance on a wide range of datasets. Due to space limits, we only consider evaluation datasets and leave the art of training data construction to future work.

{\small
\bibliographystyle{ieee_fullname}
\bibliography{bear}
}

\appendix
\section{Datasets}
\subsection{Sports datasets}
Sports-related videos are one of the most numerous videos in human visual records. Sports videos could contain a variety of different action patterns; they can be as simple as jogging and jumping, or as more complicated as some professional actions like cross-over in basketball games, all of which could be qualified learning samples for various action recognition models. In our sports datasets section, we selected three representative datasets: Sports1M~\cite{karpathy2014large}, MOD20~\cite{perera2020multiviewpoint}, and FineGym\cite{shao2020finegym}. \textbf{Sports1M} is one of the biggest sports video datasets in the vision community, which contains 487 categories and has been well-annotated. Considering the fact that some of its URLs are no longer available as well as its huge amount (the original version contains over 1 million videos), we construct a mini version of it, which only includes 50 samples per class, 40 for training, and 10 for testing. \textbf{MOD20} is a multi-viewpoint outdoor dataset collected from both YouTube videos and a drone camera, which alleviates the dataset scarcity in terms of viewpoint. Specifically, there are totally four types of views in MOD20, three of which are above the person and the fourth is an elevation view. Furthermore, these two datasets only contain coarse sports categories, ignoring the analysis of sub-actions within a sports event, which may weaken the scope of our benchmark. Considering this, we include a recently released fine-grained dataset, \textbf{FineGym}. In practice, we use one of its versions -- FineGym99, which is composed of 99 fine-grained gymnastic actions from top-level world competitions.

\subsection{Daily datasets}
We construct our daily action datasets based on two rules: abundant first-person-video data and diverse activity categories. Studies on egocentric video analysis could help us to forge an in-depth understanding of the interactions between humans and surroundings, which is essential for many cutting-edge AI technologies such as embodied AI. \textbf{CharadesEgo}~\cite{sigurdsson2018charades} is a large-scale dataset with paired first- and third-person videos to facilitate the investigation of the intrinsic correspondence between different views for the same action. In our benchmark, we choose to only carry out the evaluation using its 1st person part, since we do not focus on the correlations between two different views. Based on its official temporal annotations, we manage to segment the original videos into 43,594 short clips. \textbf{HACS}~\cite{zhao2019hacs}, human action clips and segments, is a large-scale dataset for both action recognition and temporal action localization. We only leverage HACS clips for our action recognition studies; likewise, considering its scale, we randomly sample 50 videos per class to form  mini-HACS, 40 for training and 10 for testing. In addition, we also include two more self-collected datasets based on videos captured from real daily events participants, Toyota Smarthome~\cite{das2019toyota} and MPII Cooking~\cite{rohrbach2012database}. \textbf{Toyota Smarthome} is a 3rd view dataset containing videos from different cameras deployed in an apartment, whose subjects are 18 senior people. The videos are collected from 7 cameras in the dining room, kitchen, and living room. We use the cross-subject train-test split in our evaluation, i.e., the training data are from 11 subjects and the rest are utilized for testing. \textbf{MPII Cooking} dataset is a fine-grained cooking activity dataset, which is originally built for action detection. In our benchmark, we obtain action clips containing one action label given the official temporal annotation. The raw videos are recorded based on 12 participants, and we use videos from 10 subjects as the training set.

\subsection{Anomaly datasets}
Action analysis for anomaly or crime-related videos is an important real-world application, and one of the objectives of our proposed benchmark is to provide practical guidance for the application scenario of human action recognition models. Hence, we build the anomaly track with three representative datasets to evaluate the performance of various models in such a realistic case. \textbf{UCF-Crime}~\cite{sultani2018real} is a challenging anomaly video dataset collected from surveillance cameras. We select 12 human-related crime categories from its original 14-class recognition version as we only focus on human actions. \textbf{XD-violence}~\cite{wu2020not} is another video anomaly dataset that includes data from various sources such as action movies, sports videos, and CCTV cameras. This results in a more extensive collection of video samples, enriching our anomaly datasets track. Specifically, the original XD-violence has a multi-label text set with temporal annotation, according to which we segment the test videos into single-label clips. We also include a recently released fall detection dataset \textbf{MUVIM}~\cite{denkovski2022multi} (Multi Visual Modality Fall Detection Dataset), which consists of visual data from multiple sensors: infrared, depth, RGB, and thermal cameras. Considering the data consistency, we only utilize their RGB version.

\subsection{Instructional datasets}
Instructional videos are captured in order to guide people to accomplish particular operations, e.g., assembling some objects with the components, operating a clinical surgery, or other necessary tasks which require additional knowledge. Accurate video analysis for such instructional videos is an important and irreplaceable phase for many practical applications, such as intelligent robots for industrial or medical usage. \textbf{COIN}~\cite{tang2019coin} dataset is a large-scale dataset built for comprehensive instructional video analysis based on videos collected from YouTube. It consists of 180 tasks in 12 different domains related to tasks about daily living, e.g., 'change the car tire' and 'replace the door knob'. \textbf{InHARD}~\cite{dallel2020inhard}, Industrial Human Action Recognition Dataset, is collected in a human-robot collaboration scenario. 16 distinct subjects are invited to finish an assembly task with the guidance of a robotic arm. The classes contain the specific actions during this operation, such as 'put down measuring rod' and 'put down component'. Similarly, we include another instructional dataset \textbf{MECCANO}~\cite{ragusa2021meccano}, which is also related to an assembly operation but collected with wearable cameras. The target task is to build a toy motorbike given all the components and the booklet, and the whole assembly process is precisely divided into 61 action steps. To cover scenarios as much as possible, we add two medical instructional datasets \textbf{MISAW}~\cite{huaulme2021micro} (Micro-Surgical Anastomose Workflow) and \textbf{PETRAW}~\cite{huaulme2022peg} (PEg TRAnsfer Workflow). Both two datasets are collected in simulated environments. The whole process is constructed by step-wise professional clinical operations, such as 'suturing' and 'knot tying'. Both two datasets provide frame-wise annotation in terms of phase, step, and action labels for the left hand and right hand and MISAW additionally provides the target and tool annotations. To generate segment-level action annotation, for MISAW, we view the action label of the left hand and the corresponding target as a whole, when any of them changes, we change the segment annotation. For instance, if the annotation of the action and the target for the current frame are 'Hold' and 'Left artificial vessel', we annotate the segment where it belongs as 'Hold Left artificial vessel'; if the annotation changes in the next frame into 'Catch' and 'Needle', we start a new segment and annotate it as 'Catch Needle' until the next change. Similarly, we segment PETRAW videos based on the change of the left-hand action.

\subsection{Gesture datasets}
Gesture recognition is critical for application in human-computer interfaces and has become an appealing topic in recent years. In this track, we view all datasets whose data semantic originated from symbols constructed by human body parts as the generalized gesture datasets, e.g., gestures, sign language, and other body language or pose. \textbf{Jester}~\cite{materzynska2019jester} is collected from 1,376 actors based on 27 gesture classes. The categories contained in Jester include gestures that usually appear in interactions between humans and some smart devices, such as ``Zoom in with two fingers". \textbf{WLAS}L~\cite{li2020word}, short for World-Level American Sign Language, is built for sign language understanding, which could make progress for the communications of the blind and deaf. Its original version contains 2000 categories of common sign language; in our benchmark, we use its subset WLASL100 for our evaluation. Besides, \textbf{UAV Human} dataset~\cite{li2021uav} contains videos captured from unmanned aerial vehicles, which also includes body sign language, e.g., the victory sign posed by two arms. For completeness, we also keep other regular classes of this dataset.


\section{Training Details}
\label{appendix training details}
\subsection{Supervised pre-training}
We pre-train all of the 6 models on Kinetics400. Specifically, the total training epochs for CNNs and transformers are 50 and 30, respectively. For VideoSwin, we utilize AdamW as the optimizer, while we use SGD for the rest of the models. In testing, we adopt single-view evaluation for all datasets. The frame sampling strategy is sampling 8 frames in total and sampling 1 frame per 16 frames. The weight decay is 1e-4 and momentum is 0.9 for all models.
The complete training details are shown in Table~\ref{sup training setting}.

\begin{table}[t]
\centering
\caption{The training details of our supervised pre-training.}
\scalebox{0.6}{
\begin{tabular}{c|cccccc}
\hline
HyperParams               & TSN        & TSM          & I3D        & NL         & TimeSformer        & VideoSwin\\\hline
Batch Size                 & 64        & 64           & 64         & 64         & 64                 &  64 \\ 
lr                        & 0.05       & 0.05         & 0.01        & 0.01       & 5e-3               &  5e-4 \\ 
lr policy                 & StepLR   & StepLR     & StepLR   & StepLR   & StepLR           &  CosineLR \\ 
lr step                 & [20, 40]    & [20, 40]      & [20, 40]   & [20, 40]   & [10, 20]           & /  \\ 
$\#$ Epoch                & 50         & 50           & 50         & 50         & 30                 &  30 \\ 
$\#$ WarmUp               &    /      &    /        & 10         & 10         & /                 &  2.5 \\ 
Optimizer                 & SGD        & SGD          & SGD        & SGD        & SGD              &  AdamW \\ \hline
\end{tabular}}
\label{sup training setting}
\end{table}

\subsection{Self-supervised pre-training}
We utilize {$\rho$}MoCo~\cite{rhomoco}, which is an extension of MoCo~\cite{he2020momentum} in the video domain, as our self-supervised pre-training method. Specifically, {$\rho$}MoCo, where {$\rho$} stands for the number of temporal views for contrastive learning, aims at learning an encoder that could generate invariant features for different clips of the same video. In our pre-training, we set {$\rho$}=2. The weight decay is 1e-4 and the momentum is 0.9. The complete training details are presented in Table~\ref{ssl training setting}.

\begin{table}[t]
\centering
\caption{The training details of our self-supervised pre-training.}
\scalebox{0.6}{
\begin{tabular}{c|cccccc}
\hline
HyperParams               & TSN        & TSM     & I3D        & NL    & TimeSformer        & VideoSwin\\\hline
Batch Size                 & 64      &    64      & 64         & 64   & 64                 &  64 \\ 
lr                        & 0.05      &    0.05      & 0.1         & 0.01   & 5e-4        &  5e-4 \\ 
lr policy                 & CosineLR      &    CosineLR      & CosineLR         & CosineLR   & CosineLR      &  CosineLR \\ 
$\#$ Epochs              & 50      &    50      & 50         & 50   & 50                 &  50 \\ 
$\#$ WarmUp              & 10      &    10      & 10         & 10   & 10                 &  10 \\ 
Optimizer               & SGD      &    SGD      & SGD         & SGD   & AdamW                 &  AdamW \\ \hline
\end{tabular}}
\label{ssl training setting}
\end{table}

\subsection{Standard Finetuning}
Similarly, we adopt the same training settings in supervised pre-training for our standard finetuning experiments except for the frame interval. Considering the average frame number of the clips could vary across datasets, we select different frame intervals for each dataset according to its average frame number. The details are shown in Table~\ref{frame interval}. Other training settings are the same as standard finetuning.

\begin{table}[h]
\centering
\caption{Sampling frame interval of different datasets.}
\scalebox{1.}{
\setlength{\tabcolsep}{1mm}{
\begin{tabular}{c|cc}
\hline
Dataset               & Avg. $\#$ Frames        & Frame Interval \\\hline
XD-Violence           &    517            &    16       \\ 
UCF-Crime             &  402            &   16     \\
MUVIM                 &   194            &    16      \\ \hline
WLASL100 & 68 & 8 \\
Jester & 36 & 4 \\
UAV Human & 133 & 16 \\\hline
CharadesEgo & 298 & 16 \\
Toyota Smarthome & 248 & 16 \\
MPII Cooking & 78&  8\\\hline
Mini-Sports1M & 711 & 16\\
FineGym99 &74 & 8 \\
MOD20 & 216 & 16\\\hline
COIN & 160 & 16 \\
MECCANO & 21 & 2 \\
INHARD & 71 & 8\\
PETRAW & 65  & 8 \\
MISAW & 117 & 16 \\\hline
\end{tabular}}}
\label{frame interval}
\end{table}

\subsection{Few-shot learning}
Since different training sample choices can have a large impact on few-shot learning, we randomly generate 3 training splits for all datasets and report the average performance to reduce such variation. And there are some datasets that contain categories that only have 1 or 2 samples, we ignore the training number shortage. The training setting is the same as the standard finetuning.

\section{Standard finetuning}
\label{appendix full}
We showcase the complete finetuning results in Table~\ref{tab:sup res} and Table~\ref{tab: ssl res}, which include both top-1 and top-5 accuracy based on both supervised pre-training and self-supervised pre-training. We only provide top-1 results in the main paper. Besides, we also attach existing SoTA results, if any, for each dataset in BEAR for reference in Table~\ref{tab:sup res}. For MUVIM, MPII-Cooking, and InHARD, there are no existing reported results for action recognition. For XD-Violence and CharadesEGO, the original data contains multiple labels for one long video, but we segment them into single-label clips via their temporal annotation. Besides, the SoTA results of Mini-HACS and Mini-Sports1M are actually from the complete version, we only include them for reference.


\begin{table*}[!t]
    \caption{Top-1 and top-5 accuracy of finetuning based on supervised pre-training and SoTA results for each dataset.}\label{tab:sup res}
    \centering
    \scalebox{0.9}{
    \begin{tabular}{c|cccccc|c}
          \hline
            \textbf{Dataset}          &  \textbf{TSN} &\textbf{TSM} & \textbf{I3D} & \textbf{NL} & \textbf{TimeSFormer} & \textbf{VideoSwin}  & SoTA   \\ \hline
         \textbf{XD-Violence}         & \textbf{85.54/NA}         & 82.96/NA       & 79.93/NA       & 79.91/NA       & 82.51/NA  & 82.40/NA  & $-$  \\ 
         \textbf{UCF-Crime}           & 35.42/77.78         & \textbf{42.36/79.17}       & 31.94/77.08         &  34.03/81.94      & 36.11/76.39  & 34.72/77.78 & 28.4\cite{sultani2018real}  \\ 
         \textbf{MUVIM}               & 79.30/NA         & \textbf{100/NA}         &  97.80/NA      &  98.68/NA      & 94.71/NA        &  \textbf{100.00/NA}   & $-$   \\ \hline
         \textbf{WLASL}               & 29.63/62.96        & 43.98/77.31       &   49.07/78.70      &  \textbf{52.31/78.24}    & 37.96/73.61    & 45.37/75.46  & 83.30\cite{hu2021signbert} \\ 
         \textbf{Jester}              & 86.31/99.66         & \textbf{95.21/99.77}       & 92.99/99.68        &  93.49/99.66      & 93.42/99.61    & 94.27/99.68  & 98.15\cite{truong2022direcformer}\\ 
         \textbf{UAV-Human}           & 27.89/50.82         & \textbf{38.84/61.47}       &  33.49/59.74       & 33.03/54.21         & 28.93/51.14     & 38.66/61.42 & 37.98~\cite{cheng2020skeleton} \\ \hline
         \textbf{CharadesEGO}         & 8.26/30.38          & 8.11/29.49        &  6.13/21.86        & 6.42/22.03         & \textbf{8.58/29.96}  & 8.55/29.86 & $-$ \\ 
         \textbf{Toyota Smarthome}    & 74.73/95.95         & \textbf{82.22/96.74}       & 79.51/95.60        & 76.86/94.52      & 69.21/93.30   & 79.88/97.13 & 71.0~\cite{das2021vpn++}  \\ 
         \textbf{Mini-HACS}           & 84.69/98.04         & 80.87/96.48       &77.74/95.17         &  79.51/95.17      & 79.81/96.48    & \textbf{84.94/97.58}  & 95.5~\cite{li2022uniformerv2}   \\ 
         \textbf{MPII Cooking}        & 38.39/71.17         & 46.74/74.96       & \textbf{48.71/74.05}         &    42.19/70.11    & 40.97/68.44  & 46.59/80.88  & $-$ \\ \hline
         \textbf{Mini-Sports1M}       & 54.11/80.74         & 50.06/76.57       &46.90/72.85         & 46.16/72.77          & 51.79/77.15   & \textbf{55.34/80.18} & 75.5~\cite{tran2019video}    \\ 
         \textbf{FineGym}             & 63.73/94.60         & \textbf{80.95/98.49}       &   72.00/96.14         & 71.21/95.94         & 63.92/93.88    & 65.02/92.89 & 80.4~\cite{shao2020finegym}  \\ 
         \textbf{MOD20}               & \textbf{98.30/99.86}         & 96.75/100       &  96.61/100      &  96.18/100        & 94.06/99.72   & 92.64/99.72   & 74.0~\cite{perera2020multiviewpoint} \\ \hline
         \textbf{COIN}                & 81.15/96.19         & 78.49/95.24       & 73.79/92.58        & 74.30/92.07        & \textbf{82.99/96.70}   & 76.27/93.53  & 88.02~\cite{tang2019coin}  \\ 
         \textbf{MECCANO}             & \textbf{41.06/75.20}         & 39.28/70.88       & 36.88/67.45        & 36.13/66.63          & 40.95/75.17  & 38.89/72.19  & 42.85~\cite{ragusa2021meccano}     \\ 
         \textbf{InHARD}              & 84.39/98.99         & \textbf{88.08/98.99}       &  82.06/98.63      & 86.31/98.99       & 85.16/99.23  & 87.60/99.11  & $-$   \\ 
         \textbf{PETRAW}              & 94.30/99.92         & 95.72/99.97       & 94.84/99.92        & 94.54/99.85      & 94.30/99.87    & \textbf{96.43/99.90}  & 88.51~\cite{huaulme2022peg}  \\ 
         \textbf{MISAW}               & 61.44/94.34         & \textbf{75.16/97.17}       &  68.19/96.08     & 64.27/95.64        & 71.46/96.65      & 69.06/97.17  & 63.4~\cite{huaulme2021micro}  \\ \hline
    \end{tabular}
    }
\end{table*}

\begin{table*}[!t]
    \caption{Top-1 and top-5 accuracy of finetuning based on self-supervised pre-training.}\label{tab: ssl res}
    \centering
    \scalebox{1}{
    \begin{tabular}{c|cccccc}
          \hline
        \textbf{Dataset}               & \textbf{TSN} &\textbf{TSM} & \textbf{I3D} & \textbf{NL} & \textbf{TimeSFormer} & \textbf{VideoSwin}   \\ \hline
         \textbf{XD-Violence}           & 80.49/NA        & \textbf{81.73/NA}       & 80.38/NA        & 80.94/NA       & 77.47/NA                & 77.91/NA     \\ 
         \textbf{UCF-Crime}             & \textbf{37.50/83.33}        & 35.42/81.94       & 34.03/80.56        &  34.72/83.33      &   36.11/77.78             & 34.03/80.56               \\ 
         \textbf{MUVIM}                     & 99.12/NA        & \textbf{100/NA}         & 66.96/NA        & 66.96/NA       & 99.12/NA    & \textbf{100/NA}  \\ \hline
         \textbf{WLASL}                    & 27.01/50.22        & 27.78/53.70       & 29.17/65.74        & \textbf{30.56/62.50}       &   25.56/59.44             & 28.24/65.28            \\ 
         \textbf{Jester}                     & 83.22/99.23        & \textbf{95.32/99.78}       & 87.23/99.46        &  93.89/99.68      & 90.33/99.41                & 90.18/99.40              \\ 
         \textbf{UAV-Human}             & 15.70/35.89        & 30.75/55.07       & 31.95/56.69        &26.28/51.09        & 21.02/44.28               & \textbf{35.12/59.47}          \\ \hline
         \textbf{CharadesEGO}          & 6.29/23.52         & 6.59/23.81        & 6.24/22.25         & 6.31/22.74        & 7.59/27.81            & \textbf{7.65/27.41}             \\ 
         \textbf{Toyota Smarthome}           & 68.71/91.61        & \textbf{81.34/96.63}       & 77.82/95.53        & 76.16/93.58       & 61.64/91.44                  & 80.18/97.00              \\ 
         \textbf{Mini-HACS}                  & 64.60/90.38        & 63.24/90.38       & 70.24/92.20        & 60.57/86.05       & 73.92/95.73                  &  \textbf{75.58/95.62}           \\ 
         \textbf{MPII Cooking}               & 34.45/66.16        & \textbf{50.08/75.42}       & 42.79/72.08        & 40.36/70.56       & 35.81/64.34               & 47.19/76.33            \\ \hline
         \textbf{Mini-Sports1M}              & 43.02/71.23        & 43.59/70.86       & 46.28/73.12        & 45.56/72.07       & 44.60/72.44                & \textbf{47.60/73.88}              \\ 
         \textbf{FineGym}                     & 54.62/91.21        & \textbf{75.87/97.84}       & 69.62/95.57        & 68.79/95.70       & 47.60/85.76                & 58.94/92.56              \\ 
         \textbf{MOD20}                     & 91.23/98.87        & 92.08/99.29       & 91.94/99.58        & 92.08/99.58       & 90.81/99.43               & \textbf{92.36/99.58}             \\ \hline
         \textbf{COIN}                     & 61.48/88.52        & 64.53/89.85       & 71.57/92.20        & \textbf{72.78/92.58}       & 67.64/89.97                  & 68.78/89.28              \\ 
         \textbf{MECCANO}                     & 32.34/65.92         & 35.10/65.99        & 34.86/66.99        & 33.62/66.03        & 33.30/67.87              & \textbf{37.80/72.30}              \\ 
         \textbf{InHARD}                   & 75.63/97.68          & \textbf{87.66/99.46}         & 82.54/98.87        & 80.81/98.63         & 71.28/97.38           & 80.10/98.87              \\ 
         \textbf{PETRAW}                  & 93.18/99.87        & \textbf{95.51/99.92}       & 95.02/99.92        & 94.38/99.87       & 85.56/99.38                  &  91.46/99.90             \\ 
         \textbf{MISAW}                    & 59.04/90.20        & \textbf{73.64/97.82}       & 70.37/96.73        & 64.27/94.55       & 60.78/97.39                  &  68.85/97.39             \\ \hline
    \end{tabular}}
\end{table*}

\section{Few-shot learning}
\label{appendix few}
We showcase the complete few-shot learning results in Table~\ref{tab:sup res1}, Table~\ref{tab:sup res2}, Table~\ref{tab:sup res3}, Table~\ref{tab:ssl res1}, Table~\ref{tab:ssl res2} and Table~\ref{tab:ssl res3},  which include both top-1 and top-5 accuracy for all the 3 training split based on both supervised pre-training and self-supervised pre-training.

\begin{sidewaystable}[]
\caption{Top-1 and top-5 accuracy of few-shot learning based on supervised pre-training on training split1. NL means NonLocal network and TSF means TimeSformer.}\label{tab:sup res1}
\centering
\scalebox{0.5}{
\begin{tabular}{c|c|cccccccccccccccccc}
            & $\#$Shot & XD-Violence & UCF-Crime   & MUVIM    & WLASL       & Jester      & UAV Human   & CharadesEGO & Toyota Smarthome & Mini HACS   & MPII Cooking & Mini Sports1M & FineGym     & MOD20       & COIN        & MECCANO    & InHARD      & PETRAW      & MISAW       \\\hline
\multirow{4}*{\textbf{TSN}}  & 16      & 65.81/NA    & 38.89/84.03 & 90.75/NA & 35.65/64.35 & 31.16/68.39 & 17.04/40.07 & 3.80/15.69  & 39.79/80.03      & 79.86/96.17 & 14.87/43.10  & 45.52/74.52   & 9.87/34.41  & 94.34/99.86 & 77.28/94.86 & 3.68/14.28 & 16.69/60.13 & 33.63/92.82 & 28.54/74.07 \\
            & 8 & 61.21/NA    & 33.33/75.69 & 53.30/NA & 15.74/43.52 & 20.91/52.08 & 9.56/26.89  & 2.75/12.53  & 27.37/70.75      & 76.08/95.52 & 15.02/39.00  & 39.86/69.75   & 6.24/25.62  & 95.19/99.86 & 71.07/92.45 & 1.77/8.68  & 15.32/52.44 & 27.12/83.38 & 19.17/58.17 \\
            & 4 & 58.86/NA    & 24.31/74.31 & 41.41/NA & 5.09/15.28  & 12.06/38.57 & 4.34/14.45  & 2.67/10.24  & 14.38/46.36      & 70.64/92.09 & 11.53/30.96  & 30.86/60.72   & 4.87/21.65  & 93.21/99.86 & 62.88/89.66 & 1.49/8.01  & 15.38/47.79 & 21.73/77.42 & 13.94/48.58 \\
            & 2 & 57.96/NA    & 20.83/68.06 & 37.44/NA & 1.39/6.94   & 9.54/34.68  & 1.82/6.67   & 2.37/8.11   & 10.79/37.59      & 62.59/89.63 & 4.86/29.59   & 22.85/50.76   & 22.85/50.76 & 85.71/99.72 & 51.52/83.06 & 0.74/7.01  & 10.67/41.84 & 23.99/78.68 & 10.02/35.95 \\\hline
\multirow{4}*{\textbf{TSM}} & 16        & 65.47/NA    & 42.36/79.86 & 99.56/NA & 47.69/78.24 & 36.86/76.49 & 20.20/43.08 & 4.26/16.44  & 46.90/83.23      & 76.38/94.41 & 21.40/50.23  & 42.94/70.00   & 15.43/45.98 & 93.92/99.72 & 72.08/92.20 & 4.71/16.12 & 22.47/73.78 & 61.24/97.82 & 37.47/76.69 \\
            & 8 & 59.19/NA    & 29.17/79.17 & 68.28/NA & 26.85/58.80 & 21.82/55.45 & 12.48/31/51 & 2.82/11.47  & 38.14/75.06      & 74.57/94.21 & 19.73/40.67  & 36.00/64.52   & 8.90/32.29  & 94.06/99.72 & 66.94/88.90 & 3.75/15.20 & 22.05/69.55 & 51.67/93.79 & 29.19/76.91 \\
            & 4 & 54.60/NA    & 25.00/70.83 & 64.32/NA & 9.72/25.93  & 12.36/37.49 & 6.08/19.36  & 2.31/9.48   & 24.15/56.99      & 67.52/92.30 & 8.80/24.89   & 27.76/56.22   & 6.62/25.41  & 88.26/99.58 & 57.17/85.74 & 2.44/10.52 & 18.77/66.09 & 41.61/80.53 & 14.60/51.63 \\
            & 2 & 50.22/NA    & 22.92/71.53 & 61.23/NA & 4.17/12.50  & 9.92/34.89  & 2.82/9.63   & 2.30/8.77   & 13.80/43.94      & 60.67/88.52 & 7.74/31.11   & 21.58/48.28   & 6.13/21.71  & 83.31/98.59 & 45.56/76.84 & 2.90/10.17 & 16.09/58.76 & 40.46/83.79 & 9.59/36.17  \\\hline
\multirow{4}*{\textbf{I3D}} & 16        & 58.07/NA    & 33.33/80.56 & 85.46/NA & 49.07/79.63 & 35.22/74.06 & 19.13/43.10 & 2.90/11.44  & 45.32/82.74      & 76.69/95.17 & 17.91/43.55  & 39.36/67.21   & 14.48/46.29 & 94.20/100   & 66.88/90.10 & 4.18/14.74 & 29.68/77.29 & 48.36/94.07 & 39.22/83.44 \\
            & 8 & 51.57/NA    & 27.08/70.14 & 65.2/NA  & 29.63/66.67 & 25.11/60.70 & 11.69/30.16 & 2.55/10.28  & 35.93/74.32      & 74.27/93.81 & 11.53/40.52  & 34.15/61.42   & 9.44/33.60  & 94.34/100   & 61.17/86.55 & 2.30/10.24 & 21.33/62.57 & 42.18/85.99 & 21.79/67.54 \\
            & 4 & 56.17/NA    & 21.53/67.36 & 32.16/NA & 10.65/33.33 & 13.79/42.29 & 5.76/19.05  & 2.06/8.93   & 20.12/56.38      & 69.94/90.89 & 13.96/34.14  & 27.21/54.62   & 7.41/26.24  & 88.68/99.72 & 53.17/82.11 & 1.52/8.40  & 20.98/57.33 & 28.50/76.19 & 13.51/56.21 \\
            & 2 & 42.26/NA    & 18.06/66.67 & 37.00/NA & 4.17/16.20  & 10.46/34.38 & 3.02/9.83   & 2.05/8.12   & 18.28/48.09      & 62.39/88.37 & 6.22/24.58   & 20.33/34.91   & 5.67/23.13  & 84.87/98.87 & 41.12/72.97 & 2.66/12.65 & 19.01/60.85 & 32.94/79.01 & 6.75/32.68  \\\hline
\multirow{4}*{\textbf{NL}} & 16   & 58.52/NA    & 29.17/81.25 & 93.39/NA & 51.39/83.33 & 38.16/76.95 & 18.65/41.09 & 2.93/11.09  & 42.35/78.65      & 76.28/94.01 & 19.58/47.34  & 38.97/66.02   & 15.03/44.06 & 90.38/99.86 & 68.46/90.55 & 5.32/16.43 & 25.27/72.94 & 56.77/96.05 & 33.12/69.93 \\
            & 8 & 54.04/NA    & 25.00/72.22 & 52.42/NA & 32.87/68.52 & 26.84/63.62 & 11.12/29.52 & 2.24/9.50   & 34.68/70.02      & 74.67/93.10 & 17.45/42.64  & 33.31/60.99   & 9.02/32.59  & 90.38/100   & 60.47/87.06 & 5.01/16.10 & 20.56/65.38 & 46.79/90.58 & 22.44/65.80 \\
            & 4 & 51.01/NA    & 17.36/67.36 & 54.19/NA & 9.72/34.72  & 15.29/44.98 & 5.53/17.49  & 2.10/8.73   & 20.80/55.02      & 69.03/91.59 & 10.77/29.29  & 26.51/54.07   & 6.90/25.57  & 85.15/99.86 & 52.60/81.35 & 4.53/15.69 & 21.22/61.08 & 35.89/78.32 & 11.33/49.02 \\
            & 2 & 44.39/NA    & 17.36/64.58 & 45.37/NA & 4.63/15.28  & 10.70/35.80 & 2.85/8.75   & 1.87/7/46   & 14.23/46.88      & 60.83/88.22 & 9.10/30.35   & 20.18/45.77   & 5.65/22.38  & 81.19/99.01 & 41.05/73.35 & 4.29/16.33 & 13.83/48.03 & 34.12/79.30 & 7.41/28.76  \\\hline
\multirow{4}*{\textbf{TSF}} & 16 & 67.71/NA    & 32.64/77.78 & 62.56/NA & 40.74/72.69 & 30.47/68.33 & 14.19/32.95 & 3.80/14.36  & 33.37/74.01      & 74.67/94.66 & 21.09/48.56  & 45.05/72.61   & 11.84/39.33 & 90.81/99.92 & 79.19/95.11 & 4.50/12.22 & 14.78/53.81 & 44.54/94.20 & 38.56/78.43 \\
            & 8 & 60.76/NA    & 31.25/68.75 & 57.27/NA & 22.22/55.09 & 13.53/41.70 & 6.51/19.11  & 2.99/10.45  & 18.74/52.59      & 67.72/91.94 & 13.66/35.81  & 39.88/67.60   & 7.63/28.89  & 88.83/99.86 & 75.63/93.15 & 2.27/8.25  & 12.34/56.44 & 32.84/82.35 & 14.60/47.71 \\
            & 4 & 53.48/NA    & 23.61/58.33 & 38.33/NA & 7.41/21.30  & 8.58/31.70  & 2.76/9.95   & 2.45/8.69   & 9.96/34.73       & 60.07/86.81 & 7.74/27.77   & 31.70/59.98   & 5.89/20.78  & 84.44/99.29 & 68.40/90.23 & 1.66/7.05  & 12.99/53.28 & 32.04/82.94 & 14.16/44.88 \\
            & 2 & 47.65/NA    & 15.28/68.06 & 48.90/NA & 2.78/6.02   & 6.66/27.00  & 1.37/5.21   & 2.98/7.59   & 11.03/35.21      & 49.65/76.89 & 5.31/33.69   & 22.59/48.81   & 4.49/20.14  & 72.56/97.31 & 51.90/82.49 & 1.56/7.62  & 9.83/44.70  & 31.94/73.91 & 16.12/42.70 \\\hline
\multirow{4}*{\textbf{Swin}} & 16  & 59.30/NA    & 34.03/76.39 & 97.36/NA & 43.06/75.93 & 41.89/83.89 & 22.74/47.72 & 3.47/15.32  & 46.94/87.43      & 81.22/96.93 & 21.70/51.59  & 44.35/70.92   & 11.79/39.44 & 82.89/98.16 & 65.23/89.15 & 4.68/15.41 & 23.96/68.00 & 37.20/87.53 & 30.50/71.24 \\
            & 8 & 60.65/NA    & 31.94/72.92 & 65.20/NA & 31.02/61.11 & 30.32/70.60 & 15.59/37.57 & 2.83/10.94  & 36.96/77.21      & 76.54/96.02 & 14.26/38.54  & 37.68/66.18   & 8.38/31.63  & 77.79/97.17 & 60.22/83.38 & 2.90/12.43 & 18.41/53.16 & 25.99/79.71 & 27.67/70.15 \\
            & 4 & 57.96/NA    & 14.58/65.97 & 48.90/NA & 10.19/30.56 & 16.36/46.92 & 8.68/24.70  & 2.47/9.24   & 28.18/65.23      & 70.85/93.71 & 7.59/29.14   & 30.80/58.77   & 6.73/24.00  & 72.14/96.89 & 46.13/73.67 & 1.59/9.67  & 15.61/59.65 & 27.91/73.65 & 15.47/45.21 \\
            & 2 & 43.39/NA    & 19.94/72.22 & 67.84/NA & 2.31/10.19  & 15.04/45.65 & 4.05/13.30  & 1.84/7.93   & 16.29/44.14      & 63.34/90.43 & 6.68/28.38   & 22.73/48.81   & 5.38/23.52  & 72.70/95.62 & 37.31/64.40 & 1.56/11.97 & 7.51/39.81  & 23.40/77.63 & 7.63/29.85 \\\hline
\end{tabular}}
\end{sidewaystable}

\begin{sidewaystable}[]
\caption{Top-1 and top-5 accuracy of few-shot learning based on self-supervised pre-training on training split1. NL means NonLocal network and TSF means TimeSformer.}\label{tab:ssl res1}
\centering
\scalebox{0.5}{
\begin{tabular}{c|c|cccccccccccccccccc}
     & \# Shot & XD-Violence & UCF-Crime   & MUVIM    & WLASL       & Jester      & UAV Human  & CharadesEGO & Toyota Smarthome & Mini HACS   & MPII Cooking & Mini Sports1M & FineGym     & MOD20       & COIN        & MECCANO    & InHARD      & PETRAW      & MISAW       \\\hline
\multirow{4}*{\textbf{TSN}}  & 16      & 42.94/NA    & 27.78/76.39 & 57.71/NA & 3.24/8.33   & 5.87/26.69  & 1.76/6.33  & 2.53/10.22  & 15.55/45.00      & 57.20/85.80 & 9.56/31.71   & 34.35/63.80   & 5.35/21.94  & 79.21/97.60 & 53.43/84.45 & 2.55/8.47  & 11.20/53.81 & 25.91/78.48 & 17.86/57.95 \\
     & 8       & 38.45/NA    & 27.78/70.83 & 44.05/NA & 1.85/2.78   & 4.42/21.45  & 0.75/3.74  & 1.63/6.61   & 11.91/26.06      & 51.26/82.38 & 5.92/21.40   & 26.37/56.08   & 4.90/19.38  & 73.27/97.60 & 39.40/75.00 & 1.06/6.20  & 23.90/47.32 & 13.49/67.09 & 5.01/43.57  \\
     & 4       & 32.62/NA    & 20.83/59.03 & 40.09/NA & 0.93/2.31   & 4.52/21.40  & 0.73/3.28  & 1.13/5.93   & 3.87/26.85       & 41.69/74.22 & 4.10/20.18   & 17.45/43.84   & 4.98/21.17  & 68.46/96.89 & 24.87/60.72 & 3.72/11.02 & 3.40/41.66  & 19.04/65.57 & 5.01/38.13  \\
     & 2       & 39.35/NA    & 18.06/63.89 & 42.73/NA & 0.93/2.78   & 3.44/17.89  & 0.69/3.42  & 1.10/4.57   & 6.59/25.23       & 29.81/60.83 & 0.91/13.96   & 9.73/28.36    & 3.14/14.24  & 55.16/92.50 & 13.01/35.85 & 1.88/14.28 & 12.22/51.79 & 17.55/76.37 & 4.36/13.73  \\\hline
\multirow{4}*{\textbf{TSM}}  & 16      & 58.30/NA    & 31.94/79.17 & 79.74/NA & 22.22/55.09 & 25.62/66.22 & 8.67/26.31 & 2.49/10.29  & 41.71/80.95      & 54.58/84.44 & 16.54/47.04  & 34.87/63.55   & 10.03/36.49 & 81.19/98.73 & 57.17/84.64 & 2.62/10.91 & 11.86/49.34 & 51.00/95.33 & 30.50/74.07 \\
     & 8       & 56.73/NA    & 34.03/79.86 & 68.72/NA & 6.94/24.07  & 8.16/30.74  & 2.56/9.35  & 1.60/7.95   & 28.88/69.83      & 48.44/80.21 & 9.10/32.78   & 27.31/55.24   & 6.60/23.13  & 77.09/99.01 & 47.53/78.24 & 1.70/8.25  & 20.20/57.33 & 41.46/93.12 & 15.90/60.78 \\
     & 4       & 34.30/NA    & 29.86/77.78 & 64.84/NA & 1.85/3.24   & 4.59/21.74  & 0.84/3.79  & 1.30/6.25   & 11.34/33.72      & 42.50/75.83 & 5.16/20.94   & 20.21/46.18   & 5.63/20.29  & 68.60/96.61 & 34.84/68.02 & 2.98/9.39  & 6.32/29.80  & 20.55/83.02 & 9.59/37.47  \\
     & 2       & 38.45/NA    & 15.97/61.81 & 58.15/NA & 0.93/2.31   & 4.32/21.33  & 0.76/3.59  & 1.24/4.82   & 7.18/33.00       & 34.04/67.37 & 4.10/16.08   & 14.05/35.95   & 4.54/17.48  & 65.91/96.61 & 24.24/55.84 & 1.91/8.11  & 9.77/31.94  & 18.91/80.37 & 12.64/35.29 \\\hline
\multirow{4}*{\textbf{I3D}}  & 16      & 56.73/NA    & 28.47/77.78 & 93.27/NA & 31.48/60.19 & 29.49/72.09 & 7.04/22.01 & 2.71/11.28  & 40.20/77.93      & 62.03/89.58 & 13.35/41.27  & 37.17/65.20   & 9.37/32.98  & 79.35/98.44 & 64.85/88.90 & 3.01/10.38 & 10.79/39.15 & 46.33/96.33 & 30.72/71.68 \\
     & 8       & 54.04/NA    & 29.86/75.00 & 51.68/NA & 10.65/34.26 & 10.72/37.28 & 1.77/6.59  & 2.09/8.65   & 33.13/69.61      & 56.24/86.86 & 9.41/33.08   & 29.94/57.74   & 5.98/24.10  & 77.37/98.44 & 57.87/84.39 & 1.63/8.43  & 13.29/29.80 & 41.20/90.12 & 18.95/63.62 \\
     & 4       & 49.55/NA    & 26.39/70.83 & 46.21/NA & 2.78/5.56   & 5.77/24.25  & 0.95/3.75  & 1.65/6.66   & 15.33/48.00      & 47.73/78.65 & 4.25/20.18   & 22.40/48.87   & 5.19/18.84  & 68.88/97.74 & 42.45/76.02 & 1.20/6.87  & 20.50/68.89 & 27.35/77.42 & 9.37/37.04  \\
     & 2       & 43.05/NA    & 24.31/69.44 & 45.88/NA & 1.39/1.85   & 4.51/21.64  & 0.81/3.59  & 1.25/4.94   & 8.04/33.92       & 29.76/57.96 & 4.55/21.70   & 12.01/33.18   & 4.68/15.78  & 60.96/93.21 & 24.81/57.17 & 1.31/8.04  & 10.31/44.40 & 33.30/84.09 & 8.28/33.55  \\\hline
\multirow{4}*{\textbf{NL}}   & 16      & 55.83/NA    & 34.72/78.47 & 95.59/NA & 30.56/60.65 & 28.75/70.20 & 5.71/19.00 & 2.56/10.64  & 40.36/79.02      & 60.17/87.81 & 13.05/37.33  & 35.63/64.54   & 8.75/28.95  & 77.65/98.44 & 65.36/90.16 & 4.43/11.96 & 10.85/31.17 & 42.30/96.36 & 29.19/70.15 \\
     & 8       & 52.03/NA    & 27.78/75.69 & 51.54/NA & 10.19/28.70 & 11.81/38.63 & 1.42/5.18  & 1.94/7.74   & 29.28/68.14      & 52.72/84.14 & 9.10/31.87   & 29.20/57.91   & 6.37/23.65  & 74.82/97.88 & 57.17/85.96 & 2.05/9.64  & 10.61/28.31 & 35.53/83.66 & 18.30/55.77 \\
     & 4       & 49.78/NA    & 25.69/68.75 & 47.58/NA & 1.85/5.09   & 6.36/24.93  & 0.85/3.82  & 1.30/5.52   & 10.95/36.78      & 46.12/77.69 & 6.37/24.58   & 21.60/47.35   & 5.68/19.83  & 61.95/94.06 & 41.43/73.35 & 1.87/7.02  & 10.19/40.23 & 36.35/78.45 & 12.64/36.38 \\
     & 2       & 50.56/NA    & 25.00/61.81 & 44.93/NA & 0.93/1.38   & 5.69/23.54  & 0.83/3.50  & 1.06/4.78   & 5.47/33.76       & 26.74/55.34 & 11.08/23.07  & 12.67/32.81   & 4.48/15.21  & 50.92/91.80 & 26.90/56.92 & 1.22/7.94  & 8.40/31.64  & 32.58/78.81 & 11.98/30.72 \\\hline
\multirow{4}*{\textbf{TSF}}   & 16      & 56.73/NA    & 32.64/75.69 & 68.28/NA & 4.63/18.52  & 11.96/38.29 & 4.69/15.18 & 3.09/12.02  & 28.92/67.40      & 67.47/93.61 & 13.35/33.08  & 36.18/65.95   & 7.32/28.14  & 81.90/99.01 & 57.99/85.72 & 3.15/9.78  & 10.43/36.05 & 40.02/94.72 & 23.97/66.01 \\
     & 8       & 48.54/NA    & 22.22/68.75 & 66.52/NA & 2.31/10.19  & 7.34/28.14  & 1.86/6.50  & 2.25/9.26   & 20.12/48.44      & 61.13/90.53 & 8.19/24.43   & 29.10/59.69   & 5.67/21.98  & 82.32/98.44 & 52.09/81.60 & 2.05/8.78  & 10.25/45.29 & 34.68/85.09 & 7.84/39.87  \\
     & 4       & 45.40/NA    & 18.06/60.42 & 54.19/NA & 0.93/3.24   & 6.02/24.72  & 1.01/4.30  & 1.79/8.04   & 12.85/35.10      & 52.27/84.49 & 2.43/15.78   & 21.95/50.53   & 5.68/19.92  & 78.08/96.61 & 40.42/71.26 & 1.56/9.74  & 12.51/43.27 & 31.66/87.74 & 4.14/38.13  \\
     & 2       & 46.64/NA    & 17.36/61.81 & 74.89/NA & 1.39/3.24   & 5.07/22.60  & 0.85/3.74  & 1.46/5.78   & 7.49/34.20       & 40.99/71.60 & 2.28/16.69   & 15.32/37.93   & 4.32/18.25  & 67.61/92.50 & 25.63/55.77 & 1.52/8.04  & 8.58/41.95  & 32.22/85.99 & 5.88/24.84  \\\hline
\multirow{4}*{\textbf{Swin}} & 16      & 53.59/NA    & 36.81/80.56 & 92.95/NA & 16.20/43.06 & 23.30/57.25 & 9.96/28.69 & 2.88/11.29  & 43.55/83.71      & 69.59/92.40 & 18.82/47.04  & 38.21/66.10   & 11.73/37.65 & 79.92/98.73 & 61.29/87.25 & 3.4/12.54  & 13.77/47.74 & 26.04/84.86 & 28.10/65.36 \\
     & 8       & 48.88/NA    & 33.33/71.53 & 62.56/NA & 7.87/20.37  & 10.85/36.38 & 4.33/15.46 & 1.95/8.30   & 31.84/69.50      & 60.52/88.22 & 9.86/32.32   & 30.45/58.19   & 7.03/26.87  & 76.94/97.17 & 51.14/77.79 & 1.88/7.05  & 13.47/47.14 & 27.78/75.94 & 12.20/56.21 \\
     & 4       & 41.93/NA    & 24.31/72.22 & 70.48/NA & 2.78/6.02   & 6.03/24.85  & 2.01/6.52  & 1.75/6.55   & 14.08/49.29      & 51.46/80.97 & 7.59/23.98   & 22.90/48.99   & 5.97/22.78  & 69.02/95.76 & 38.64/65.86 & 1.20/5.92  & 13.05/45.83 & 32.20/76.94 & 5.45/43.57  \\
     & 2       & 34.42/NA    & 29.17/68.75 & 53.74/NA & 1.39/4.17   & 4.44/21.44  & 1.01/3.79  & 1.50/5.40   & 8.45/37.68       & 39.63/69.23 & 5.16/23.37   & 15.95/38.71   & 4.19/16.11  & 68.18/92.08 & 26.84/52.47 & 1.06/6.80  & 10.79/47.91 & 29.30/76.48 & 4.79/33.77 \\\hline
\end{tabular}}
\end{sidewaystable}

\begin{sidewaystable}[]
\caption{Top-1 and top-5 accuracy of few-shot learning based on self-supervised pre-training on training split2. NL means NonLocal network and TSF means TimeSformer.}\label{tab:sup res2}
\centering
\scalebox{0.5}{
\begin{tabular}{c|c|cccccccccccccccccc}
            & \# Shot & XD-Violence & UCF-Crime   & MUVIM    & WLASL       & Jester      & UAV Human   & CharadesEGO & Toyota Smarthome & Mini HACS   & MPII Cooking & Mini Sports1M & FineGym     & MOD20       & COIN        & MECCANO     & InHARD      & PETRAW      & MISAW       \\\hline
\multirow{4}*{\textbf{TSN}}         & 16      & 62.00/NA    & 36.11/81.25 & 74.45/NA & 32.41/63.89 & 32.66/70.59 & 17.69/40.94 & 4.57/17.65  & 33.83/76.37      & 80.16/96.88 & 24.58/61.31  & 45.28/74.62   & 16.16/49.44 & 96.32/100   & 76.33/95.05 & 13.35/42.30 & 41.84/88.08 & 40.12/94.72 & 38.34/83.44 \\
            & 8       & 63.90/NA    & 27.78/77.78 & 71.81/NA & 25.00/51.85 & 20.71/53.50 & 10.07/27.49 & 3.80/14.68  & 31.09/70.99      & 76.44/96.22 & 18.06/53.72  & 38.58/69.06   & 9.22/33.89  & 95.05/100   & 70.49/91.75 & 7.08/26.74  & 26.16/75.74 & 34.56/90.23 & 32.24/71.24 \\
            & 4       & 50.67/NA    & 24.31/75.00 & 66.52/NA & 14.35/30.56 & 10.08/35.88 & 4.02/12.67  & 2.70/10.63  & 14.38/46.36      & 70.39/93.66 & 14.87/45.98  & 31.70/61.66   & 7.41/28.41  & 93.21/100   & 62.50/89.02 & 6.41/22.81  & 24.49/64.36 & 23.14/80.25 & 18.52/52.29 \\
            & 2       & 42.83/NA    & 20.83/71.53 & 53.30/NA & 6.02/11.11  & 6.74/28.53  & 2.14/7.66   & 2.48/9.07   & 16.31/47.97      & 62.19/89.02 & 8.04/35.66   & 25.26/51.70   & 5.87/22.59  & 90.81/99.72 & 52.35/83.06 & 4.85/16.47  & 16.21/62.51 & 25.58/85.17 & 15.03/45.53 \\\hline
\multirow{4}*{\textbf{TSM}}         & 16      & 65.47/NA    & 42.36/79.86 & 99.56/NA & 47.69/78.24 & 36.86/76.49 & 20.20/43.08 & 4.26/16.44  & 46.90/83.23      & 76.38/94.41 & 21.40/50.23  & 42.94/70.00   & 15.43/45.98 & 93.92/99.72 & 72.08/92.20 & 4.71/16.12  & 22.47/73.78 & 61.24/97.82 & 37.47/76.69 \\
            & 8       & 59.19/NA    & 29.17/79.17 & 68.28/NA & 26.85/58.80 & 21.82/55.45 & 12.48/31/51 & 2.82/11.47  & 38.14/75.06      & 74.57/94.21 & 19.73/40.67  & 36.00/64.52   & 8.90/32.29  & 94.06/99.72 & 66.94/88.90 & 3.75/15.20  & 22.05/69.55 & 51.67/93.79 & 29.19/76.91 \\
            & 4       & 54.60/NA    & 25.00/70.83 & 64.32/NA & 9.72/25.93  & 12.36/37.49 & 6.08/19.36  & 2.31/9.48   & 24.15/56.99      & 67.52/92.30 & 8.80/24.89   & 27.76/56.22   & 6.62/25.41  & 88.26/99.58 & 57.17/85.74 & 2.44/10.52  & 18.77/66.09 & 41.61/80.53 & 14.60/51.63 \\
            & 2       & 50.22/NA    & 22.92/71.53 & 61.23/NA & 4.17/12.50  & 9.92/34.89  & 2.82/9.63   & 2.30/8.77   & 13.80/43.94      & 60.67/88.52 & 7.74/31.11   & 21.58/48.28   & 6.13/21.71  & 83.31/98.59 & 45.56/76.84 & 2.90/10.17  & 16.09/58.76 & 40.46/83.79 & 9.59/36.17  \\\hline
\multirow{4}*{\textbf{I3D}}         & 16      & 54.26/NA    & 34.03/75.69 & 65.64/NA & 48.15/80.09 & 37.50/76.91 & 18.48/42.31 & 3.27/12.34  & 41.86/84.24      & 77.64/95.02 & 32.02/66.46  & 39.49/67.27   & 19.97/58.71 & 94.63/100   & 67.07/90.36 & 13.71/40.45 & 55.13/90.58 & 59.08/96.74 & 47.06/86.49 \\
            & 8       & 55.94/NA    & 35.42/70.83 & 67.40/NA & 37.50/73.15 & 24.07/59.93 & 11.36/30.42 & 3.38/11.54  & 34.01/80.14      & 73.41/94.16 & 21.24/53.41  & 33.49/61.66   & 13.83/44.78 & 93.92/100   & 59.58/86.23 & 10.80/32.55 & 41.00/85.10 & 48.33/92.97 & 37.04/84.10 \\
            & 4       & 43.16/NA    & 25.69/68.06 & 53.74/NA & 19.91/42.13 & 14.93/43.29 & 5.46/17.68  & 2.47/9.47   & 23.21/59.10      & 69.39/92.35 & 14.42/44.92  & 27.13/54.66   & 8.29/30.94  & 90/95/99.86 & 53.36/82.42 & 7.12/24.09  & 27.71/71.45 & 35.15/86.45 & 27.45/65.58 \\
            & 2       & 41.26/NA    & 16.67/63.89 & 55.07/NA & 8.33/19.91  & 8.00/31.01  & 2.62/9.45   & 1.97/7.97   & 18.68/51.39      & 62.24/88.42 & 11.99/37.48  & 22.36/48.07   & 6.49/22.49  & 86.42/98.73 & 41.31/74.05 & 4.18/17.25  & 24.31/62.46 & 32.43/91.12 & 11.33/46.19 \\\hline
\multirow{4}*{\textbf{NL}}    & 16      & 56.39/NA    & 35.42/80.56 & 77.53/NA & 51.39/80.09 & 40.86/81.78 & 18.39/41.01 & 3.22/12.65  & 38.87/80.66      & 77.90/95.17 & 27.77/64.04  & 39.12/66.53   & 19.67/56.14 & 95.19/100   & 67.13/90.29 & 14.28/41.12 & 55.72/93.15 & 66.47/97.15 & 49.46/83.44 \\
            & 8       & 50.22/NA    & 25.00/72.93 & 66.96/NA & 39.35/72.69 & 25.72/63.19 & 10.72/29.09 & 3.18/10.89  & 30.04/72.21      & 73.26/94.31 & 26.10/55.39  & 32.61/60.25   & 13.17/44.59 & 92.93/100   & 62.18/86.93 & 11.32/33.46 & 41.12/84.74 & 48.59/90.94 & 31.81/74.95 \\
            & 4       & 45.74/NA    & 22.92/68.75 & 68.72/NA & 20.83/45.83 & 15.04/44.16 & 5.31/16.76  & 2.61/9.63   & 20.28/58.62      & 70.19/91.84 & 18.51/47.34  & 26.65/53.37   & 8.59/30.21  & 90.52/99.86 & 52.03/81.66 & 8.78/25.93  & 24.97/70.44 & 36.04/87.25 & 24.62/66.23 \\
            & 2       & 39.01/NA    & 21.53/62.50 & 47.58/NA & 7.41/25.00  & 8.74/34.43  & 2.85/9.49   & 2.22/7.91   & 18.76/50.29      & 60.98/87.56 & 8.8.0/43.10  & 22.55/47.66   & 5.95/22.14  & 87.27/99.72 & 42.58/73.03 & 3.75/15.80  & 20.50/59.18 & 33.94/88.30 & 15.25/48.58 \\\hline
\multirow{4}*{\textbf{TSF}} & 16      & 66.82/NA    & 31.25/78.47 & 92.95/NA & 42.59/75.93 & 45.25/84.82 & 18.75/40.89 & 4.27/14.97  & 38.69/81.37      & 81.32/96.93 & 35.05/65.55  & 46.80/74.35   & 22.40/60.81 & 95.90/99.86 & 79.12/95.49 & 14.56/42.47 & 53.04/93.33 & 64.75/97.41 & 50.33/90.41 \\
            & 8       & 60.99/NA    & 35.42/79.86 & 84.14/NA & 34.26/65.74 & 30.41/69.20 & 12.49/30.71 & 3.70/13.72  & 31.68/76.35      & 78.50/96.48 & 26.71/62.82  & 42.14/70.84   & 15.57/49.95 & 96.89/99.72 & 76.46/94.48 & 10.56/32.77 & 42.61/86.23 & 55.11/94.66 & 35.08/81.26 \\
            & 4       & 56.17/NA    & 26.39/69.44 & 81.50/NA & 17.59/44.44 & 16.93/49.71 & 6.78/20.24  & 3.33/11.27  & 24.06/61.84      & 72.96/93.61 & 17.00/50.53  & 34.95/64.66   & 11.86/41.10 & 91.23/99.86 & 70.11/91.81 & 8.18/24.48  & 21.39/65.55 & 38.25/89.92 & 23.97/62.96 \\
            & 2       & 48.32/NA    & 17.36/63.89 & 59.47/NA & 6.94/23.61  & 10.33/35.71 & 3.33/10.76  & 2.81/10.05  & 17.58/52.24      & 62.94/88.27 & 14.11/36.72  & 28.32/56.57   & 8.32/29.90  & 87.98/99.43 & 57.87/84.64 & 6.70/22.32  & 19.55/57.57 & 35.86/89.76 & 20.48/62.31 \\\hline
\multirow{4}*{\textbf{Swin}}   & 16      & 51.01/NA    & 29.86/79.17 & 92.95/NA & 43.06/75.93 & 34.23/73.22 & 13.58/34.47 & 4.10/14.87  & 43.09/82.27      & 81.92/96.83 & 34.90/67.37  & 48.97/74.58   & 18.78/55.43 & 85.71/99.15 & 61.29/85.28 & 14.35/43.18 & 44.82/84.51 & 23.55/85.63 & 36.17/84.97 \\
            & 8       & 52.58/NA    & 23.61/71.53 & 79.74/NA & 37.04/65.73 & 23.68/58.92 & 6.87/20.88  & 3.58/13.33  & 36.15/82.22      & 77.09/96.02 & 24.58/62.06  & 41.17/68.44   & 12.43/43.24 & 85.86/98.73 & 48.16/78.93 & 12.61/33.62 & 7.57/31.64  & 30.45/82.20 & 26.36/75.60 \\
            & 4       & 39.69/NA    & 22.92/65.28 & 52.42/NA & 24.54/50.00 & 13.89/43.32 & 2.89/10.30  & 2.90/11.01  & 22.07/57.30      & 71.80/93.20 & 20.79/50.23  & 33.00/60.27   & 9.30/34.81  & 76.10/97.45 & 35.98/65.80 & 6.80/23.66  & 19.13/53.58 & 27.04/79.45 & 22.66/57.08 \\
            & 2       & 33.74/NA    & 22.22/65.97 & 54.53/NA & 9.26/27.31  & 7.72/30.08  & 1.33/5.27   & 2.10/8.23   & 17.67/48.89      & 61.73/89.73 & 13.35/35.96  & 26.24/52.40   & 6.32/25.22  & 80.76/98.30 & 27.92/53.11 & 5.53/20.33  & 19.79/55.36 & 16.75/68.83 & 8.06/42.70 \\\hline
\end{tabular}}
\end{sidewaystable}

\begin{sidewaystable}[]
\caption{Top-1 and top-5 accuracy of few-shot learning based on self-supervised pre-training on training split2. NL means NonLocal network and TSF means TimeSformer.}\label{tab:ssl res2}
\centering
\scalebox{0.5}{
\begin{tabular}{c|c|cccccccccccccccccc}
     & \# Shot & XD-Violence & UCF-Crime   & MUVIM    & WLASL       & Jester      & UAV Human   & CharadesEGO & Toyota Smarthome & Mini HACS   & MPII Cooking & Mini Sports1M & FineGym     & MOD20       & COIN        & MECCANO     & InHARD      & PETRAW      & MISAW       \\\hline
\multirow{4}*{\textbf{TSN}}  & 16      & 45.74/NA    & 34.72/85.42 & 74.01/NA & 2.78/9.72   & 9.56/35.35  & 1.65/6.79   & 2.68/10.78  & 17.04/47.82      & 58.01/86.46 & 16.39/43.40  & 33.63/63.88   & 10.16/34.98 & 85.57/98.87 & 54.70/83.12 & 7.47/28.27  & 5.07/37.19  & 25.60/74.17 & 29.41/81.92 \\
     & 8       & 34.75/NA    & 20.83/61.11 & 46.70/NA & 0.46/5.09   & 4.71/21.84  & 0.83/3.79   & 1.98/8.22   & 10.09/29.23      & 51.11/81.77 & 11.99/40.21  & 25.56/55.93   & 5.92/26.24  & 80.76/99.15 & 39.97/74.56 & 7.47/23.66  & 11.14/33.08 & 18.03/79.60 & 7.63/52.51  \\
     & 4       & 36.66/NA    & 9.72/55.56  & 47.58/NA & 0.46/3.70   & 4.25/19.81  & 0.95/3.62   & 1.56/5.99   & 7.34/37.82       & 42.35/76.23 & 2.73/17.60   & 18.56/43.49   & 5.13/21.16  & 68.60/99.01 & 24.81/58.31 & 4.92/17.14  & 7.63/41.42  & 19.09/68.50 & 6.32/40.31  \\
     & 2       & 42.60/NA    & 9.72/56.25  & 43.61/NA & 1.39/2.31   & 4.40/20.63  & 0.65/3.45   & 1.05/4.22   & 3.42/26.63       & 29.76/61.33 & 3.03/8.04    & 11.25/29.61   & 5.57/17.57  & 55.16/91.23 & 15.42/38.07 & 1.45/16.65  & 7.33/47.20  & 17.34/77.09 & 7.41/27.23  \\\hline
\multirow{4}*{\textbf{TSM}}  & 16      & 56.28/NA    & 39.58/77.08 & 82.82/NA & 20.83/51.85 & 26.12/64.21 & 8.48/26.25  & 2.71/11.63  & 36.78/81.13      & 56.24/85.50 & 30.35/68.74  & 35.26/63.59   & 16.29/56.13 & 86.42/99.01 & 56.03/84.58 & 10.02/32.06 & 34.03/86.63 & 56.08/94.74 & 49.02/89.98 \\
     & 8       & 50.45/NA    & 29.86/77.78 & 68.72/NA & 8.80/30.56  & 9.66/30.81  & 2.42/9.47   & 2.52/9.78   & 32.41/74.73      & 48.64/81.87 & 20.03/59.03  & 27.52/55.24   & 10.02/38.60 & 82.32/99.15 & 46.70/77.47 & 8.08/28.48  & 16.69/47.14 & 43.82/93.02 & 35.08/81.05 \\
     & 4       & 43.61/NA    & 21.53/74.31 & 75.33/NA & 0.93/6.02   & 4.91/22.01  & 0.97/3.97   & 1.69/7.13   & 18.24/44.32      & 44.61/77.24 & 8.65/34.60   & 20.66/46.74   & 6.33/27.70  & 75.11/99.01 & 36.29/68.15 & 5.99/25.12  & 2.92/31.35  & 31.81/80.94 & 13.73/52.51 \\
     & 2       & 43.49/NA    & 11.81/63.19 & 40.09/NA & 0.46/4.63   & 4.86/22.09  & 0.63/3.34   & 1.32/5.06   & 7.27/30.94       & 35.40/67.77 & 5.46/18.51   & 14.99/37.52   & 5.41/19.41  & 66.34/97.17 & 26.97/55.77 & 4.96/24.30  & 5.30/32.18  & 18.91/73.65 & 3.70/27.89  \\\hline
\multirow{4}*{\textbf{I3D}}  & 16      & 50.34/NA    & 36.81/78.47 & 92.51/NA & 30.56/59.72 & 31.16/71.43 & 6.75/22.13  & 3.32/12.55  & 38.27/81.37      & 63.19/89.78 & 29.74/64.95  & 37.39/65.98   & 14.19/47.43 & 89.25/99.29 & 64.91/89.47 & 13.43/40.31 & 33.37/77.59 & 43.84/94.00 & 43.57/88.02 \\
     & 8       & 51.35/NA    & 27.78/76.39 & 33.04/NA & 17.14/41.67 & 12.70/38.79 & 1.28/6.06   & 2.68/10.01  & 33.85/75.28      & 56.70/86.91 & 27.62/62.52  & 30.23/58.03   & 8.54/33.90  & 85.15/99.58 & 53.17/83.57 & 9.14/30.18  & 30.63/76.16 & 39.28/90.38 & 36.82/86.71 \\
     & 4       & 41.26/NA    & 27.78/67.36 & 54.19/NA & 3.24/11.11  & 5.64/24.36  & 1.00/3.95   & 1.83/7.52   & 20.10/49.73      & 47.94/81.47 & 15.63/49.01  & 22.09/48.30   & 6.57/25.14  & 78.64/98.73 & 40.48/73.73 & 6.54/23.85  & 13.59/55.24 & 29.12/81.48 & 24.40/62.75 \\
     & 2       & 42.94/NA    & 16.67/65.97 & 61.23/NA & 1.85/6.02   & 4.69/20.46  & 0.84/3.62   & 1.56/6.13   & 5.76/29.23       & 30.61/63.14 & 10.17/29.14  & 14.99/33.98   & 4.90/19.25  & 64.92/94.63 & 28.36/58.06 & 2.13/17.07  & 8.70/38.44  & 25.04/89.12 & 11.76/44.01 \\\hline
\multirow{4}*{\textbf{NL}}   & 16      & 47.98/NA    & 35.42/77.08 & 95.15/NA & 28.70/63.43 & 31.95/73.35 & 5.11/18.07  & 3.16/12.64  & 37.33/79.35      & 61.53/89.08 & 32.47/65.10  & 36.39/65.15   & 13.16/45.17 & 87.41/99.29 & 65.04/89.28 & 15.12/42.59 & 36.23/84.21 & 39.64/87.02 & 42.48/87.58 \\
     & 8       & 48.21/NA    & 29.86/75.69 & 77.09/NA & 12.50/36.11 & 12.44/38.41 & 1.05/5.27   & 2.64/9.65   & 31.90/71.82      & 55.74/85.30 & 25.19/60.24  & 29.71/57.35   & 8.24/31.67  & 81.33/99.29 & 54.63/84.14 & 9.44/30.81  & 12.51/48.81 & 33.43/88.56 & 30.07/83.88 \\
     & 4       & 39.8/NA     & 25.00/72.92 & 63/NA    & 3.24/8.33   & 6.71/25.97  & 1.03/3.71   & 1.63/7.12   & 19.20/51.41      & 44.41/78.80 & 17.15/44.61  & 21.11/46.67   & 5.63/24.33  & 74.82/98.30 & 39.72/72.65 & 7.81/24.57  & 12.57/57.09 & 30.58/81.73 & 20.04/54.90 \\
     & 2       & 43.61/NA    & 22.22/68.75 & 54.63/NA & 0.46/4.17   & 4.82/20.92  & 0.75/3.43   & 1.36/5.34   & 5.28/28.58       & 28.05/59.02 & 6.98/32.63   & 14.02/33.84   & 4.65/19.43  & 60.82/93.49 & 28.43/56.73 & 2.38/10.12  & 4.11/34.92  & 29.84/90.56 & 6.54/36.60  \\\hline
\multirow{4}*{\textbf{TSF}}   & 16      & 54.15/NA    & 35.42/77.78 & 65.64/NA & 6.02/20.83  & 14.58/39.99 & 4.35/14.35  & 3.31/12.59  & 23.60/62.67      & 66.82/93.81 & 20.18/55.08  & 36.14/66.39   & 10.19/39.32 & 87.27/99.29 & 58.19/86.04 & 10.17/32.13 & 20.80/67.34 & 48.15/94.66 & 32.68/79.96 \\
     & 8       & 44.28/NA    & 25.00/70.14 & 67.84/NA & 3.70/12.04  & 8.02/29.24  & 1.62/5.98   & 2.94/10.64  & 20.38/52.38      & 60.37/90.28 & 13.81/39.76  & 28.79/59.24   & 7.08/29.68  & 85.71/99.01 & 48.29/80.20 & 6.55/22.17  & 12.04/55.48 & 42.36/93.46 & 17.65/64.27 \\
     & 4       & 36.88/NA    & 17.36/60.42 & 56.39/NA & 0.93/5.56   & 5.42/22.85  & 0.97/4.25   & 1.91/7.74   & 14.19/45.53      & 52.67/85.55 & 10.02/29.14  & 21.99/49.77   & 5.19/22.67  & 76.24/98.59 & 38.90/71.45 & 5.17/19.55  & 12.04/44.87 & 34.74/86.28 & 8.50/52.72  \\
     & 2       & 42.15/NA    & 15.28/58.33 & 33.92/NA & 0.46/5.56   & 4.78/21.15  & 0.97/3.71   & 1.51/6.06   & 12.39/38.06      & 39.58/69.79 & 8.04/23.52   & 16.16/38.77   & 4.54/18.14  & 57.71/89.67 & 18.59/43.40 & 6.45/24.51  & 15.14/52.21 & 33.48/85.56 & 4.79/30.07  \\\hline
\multirow{4}*{\textbf{Swin}} & 16      & 56.50/NA    & 29.86/79.17 & 92.95/NA & 20.83/49.07 & 24.55/58.80 & 13.33/34.34 & 3.56/13.02  & 36.15/80.42      & 69.03/93.30 & 31.41/67.07  & 42.67/68.85   & 15.02/47.33 & 88.25/99.29 & 61.17/85.15 & 14.91/41.09 & 48.33/87.78 & 29.86/90.12 & 36.17/84.31 \\
     & 8       & 53.36/NA    & 25.69/76.39 & 71.37/NA & 11.11/35.19 & 11.80/37.13 & 6.87/21.22  & 2.89/10.42  & 29.43/77.93      & 59.32/88.52 & 27.77/62.82  & 34.00/60.41   & 9.86/35.51  & 80.62/97.88 & 49.18/78.74 & 9.03/28.98  & 25.51/65.97 & 24.99/84.35 & 28.10/76.91 \\
     & 4       & 44.51/NA    & 24.31/70.14 & 77.09/NA & 5.56/17.59  & 6.07/24.18  & 3.13/10.28  & 1.91/7.43   & 22.57/55.84      & 50.70/81.82 & 21.24/52.05  & 25.56/50.47   & 7.22/27.19  & 74.68/96.32 & 36.99/66.12 & 5.74/19.77  & 10.61/50.66 & 15.60/79.50 & 17.43/55.99 \\
     & 2       & 44.73/NA    & 17.36/66.67 & 55.51/NA & 3.24/8.33   & 4.27/20.24  & 1.52/5.29   & 1.62/6.06   & 11.85/35.41      & 39.33/68.78 & 14.26/36.12  & 19.30/40.72   & 5.78/19.14  & 70.30/94.48 & 28.36/52.54 & 4.50/21.32  & 6.85/45.23  & 8.31/66.73  & 7.19/38.34 \\\hline
\end{tabular}}
\end{sidewaystable}

\begin{sidewaystable}[]
\caption{Top-1 and top-5 accuracy of few-shot learning based on supervised pre-training on training split3. NL means NonLocal network and TSF means TimeSformer.}\label{tab:sup res3}
\centering
\scalebox{0.5}{
\begin{tabular}{c|c|cccccccccccccccccc}
            & \# Shot & XD-Violence & UCF-Crime   & MUVIM    & WLASL       & Jester      & UAV Human   & CharadesEGO & Toyota Smarthome & Mini HACS   & MPII Cooking & Mini Sports1M & FineGym     & MOD20       & COIN        & MECCANO     & InHARD      & PETRAW      & MISAW       \\\hline
\multirow{4}*{\textbf{TSN}}         & 16      & 61.66/NA    & 38.19/75.00 & 74.89/NA & 37.04/67.59 & 16.43/39.77 & 16.43/39.77 & 4.08/15.90  & 34.38/78.04      & 79.91/96.93 & 26.56/62.06  & 45.13/74.74   & 16.89/50.97 & 95.90/100   & 75.76/95.05 & 14.28/42.47 & 42.79/87.19 & 42.89/95.74 & 35.29/76.91 \\
            & 8       & 58.52/NA    & 24.31/72.92 & 71.37/NA & 22.22/50.46 & 20.88/54.01 & 8.61/24.40  & 3.92/13.56  & 30.15/64.90      & 76.38/95.67 & 20.33/56.30  & 38.99/69.51   & 9.78/36.08  & 96.75/99.86 & 70.94/91.81 & 10.06/31.53 & 30.81/80.93 & 30.32/89.89 & 28.76/66.88 \\
            & 4       & 52.13/NA    & 22.92/66.67 & 67.84/NA & 5.09/22.22  & 11.21/37.76 & 3.74/12.74  & 3.04/11.74  & 22.66/52.22      & 69.13/93.61 & 16.54/43.85  & 30.47/61.13   & 6.41/25.25  & 93.64/99.72 & 62.82/88.58 & 7.93/24.94  & 29.56/68.71 & 20.60/82.09 & 25.53/59.04 \\
            & 2       & 38.00/NA    & 15.97/70.83 & 48.02/NA & 2.31/12.04  & 8.68/30.20  & 2.10/7.18   & 2.35/8.88   & 15.41/44.82      & 63.24/90.18 & 7.13/34.29   & 22.48/51.27   & 5.10/21.87  & 92.64/99.01 & 52.79/82.74 & 3.86/20.97  & 20.14/57.63 & 23.37/83.68 & 23.31/53.81 \\\hline
\multirow{4}*{\textbf{TSM}}         & 16      & 56.73/NA    & 34.03/75.00 & 100/NA   & 48.61/75.93 & 36.50/75.97 & 19.52/42.72 & 3.34/13.15  & 42.15/81.54      & 76.64/94.41 & 33.84/63.73  & 41.70/70.10   & 20.00/57.30 & 94.23/100   & 70.81/92.83 & 16.19/44.56 & 52.62/88.44 & 66.37/97.20 & 47.93/81.92 \\
            & 8       & 54.82/NA    & 24.31/73.61 & 99.12/NA & 31.02/65.28 & 21.50/55.82 & 11.65/29.98 & 3.00/11.51  & 31.36/68.30      & 73.41/93.61 & 29.59/58.42  & 35.79/64.64   & 11.62/40.14 & 93.35/99.86 & 63.96/88.07 & 11.02/34.93 & 37.78/82.90 & 54.77/92.38 & 33.77/70.81 \\
            & 4       & 51.46/NA    & 16.67/66.67 & 97.36/NA & 7.41/31.94  & 11.28/36.48 & 5.71/18.55  & 2.45/9.51   & 21.76/55.13      & 68.98/91.99 & 22.46/52.05  & 26.96/56.24   & 7.76/28.79  & 91.94/99.72 & 56.35/83.76 & 8.54/27.06  & 33.37/81.59 & 40.51/89.48 & 33.33/65.80 \\
            & 2       & 37.78/NA    & 14.58/60.42 & 56.39/NA & 3.24/16.20  & 9.20/30.85  & 2.54/9.68   & 1.83/7.86   & 18.13/46.40      & 59.32/87.06 & 14.87/40.67  & 20.92/46.43   & 6.35/24.21  & 89.67/99.01 & 46.51/76.52 & 5.74/21.75  & 20.44/59.24 & 37.10/87.53 & 19.83/57.52 \\\hline
\multirow{4}*{\textbf{I3D}}         & 16      & 58.30/NA    & 28.47/74.31 & 84.14/NA & 49.54/80.56 & 38.74/77.70 & 18.97/41.71 & 3.13/11.82  & 44.23/82.61      & 77.44/95.77 & 33.08/69.50  & 39.14/67.25   & 20.24/57.51 & 94.91/99.72 & 67.70/90.42 & 17.36/43.64 & 55.42/91.00 & 62.21/96.02 & 44.88/84.10 \\
            & 8       & 56.05/NA    & 25.00/71.53 & 71.37/NA & 35.65/67.59 & 25.16/62.01 & 10.40/29.54 & 2.90/10.62  & 34.29/73.27      & 74.07/95.07 & 22.91/53.26  & 33.04/62.03   & 13.49/43.27 & 94.63/99.86 & 59.96/86.10 & 12.29/33.86 & 45.23/84.98 & 46.07/93.92 & 34.42/70.15 \\
            & 4       & 52.80/NA    & 13.89/59.72 & 70.48/NA & 13.42/40.74 & 13.04/41.60 & 5.02/17.33  & 2.49/9.10   & 25.36/55.95      & 69.54/93.25 & 22.91/54.32  & 26.88/53.94   & 7.44/29.68  & 92.22/99.86 & 53.17/81.22 & 7.83/25.61  & 32.48/73.24 & 32.56/85.27 & 26.80/62.09 \\
            & 2       & 40.25/NA    & 9.72/57.64  & 48.46/NA & 3.24/18.06  & 8.83/30.15  & 2.20/9.23   & 1.83/7.84   & 17.04/45.44      & 60.78/80.27 & 14.26/33.69  & 20.60/46.08   & 6.37/23.65  & 88.54/99.01 & 42.51/73.92 & 4.82/20.30  & 21.28/59.59 & 30.30/87.94 & 19.83/44.01 \\\hline
\multirow{4}*{\textbf{NL}}    & 16      & 55.83/NA    & 23.61/71.53 & 93.83/NA & 51.85/81.02 & 39.14/79.26 & 17.92/40.41 & 3.23/11.58  & 41.27/77.45      & 76.94/94.86 & 33.84/60.55  & 38.40/65.69   & 20.19/56.78 & 95.47/100   & 68.53/89.78 & 18.31/44.95 & 56.14/93.21 & 63.19/96.79 & 42.70/81.05 \\
            & 8       & 53.14/NA    & 18.75/64.58 & 84.58/NA & 40.74/72.22 & 25.94/63.34 & 10.09/28.42 & 3.04/10.86  & 32.21/70.13      & 74.47/94.11 & 26.10/55.84  & 32.24/61.09   & 12.32/41.00 & 95.19/100   & 58.95/86.23 & 10.98/33.65 & 43.68/85.04 & 46.87/93.97 & 30.50/66.45 \\
            & 4       & 51.01/NA    & 11.11/64.58 & 87.22/NA & 12.04/38.43 & 14.62/44.05 & 4.74/16.50  & 2.47/9.72   & 25.58/57.98      & 70.59/92.65 & 22.61/53.26  & 25.59/53.31   & 7.54/29.73  & 92.22/99.58 & 52.22/81.35 & 9.92/29.58  & 39.27/78.19 & 34.94/87.17 & 28.32/61.87 \\
            & 2       & 40.36/NA    & 9.03/57.64  & 48.46/NA & 4.17/18.06  & 11.09/34.59 & 2.50/9.43   & 1.76/7.52   & 16.12/46.73      & 61.88/88.77 & 17.45/39.76  & 19.86/44.46   & 6.14/24.95  & 87.84/99.01 & 42.32/74.43 & 7.08/25.26  & 21.22/63.95 & 32.27/84.99 & 16.56/52.51 \\\hline
\multirow{4}*{\textbf{TSF}} & 16      & 63.23/NA    & 34.03/77.08 & 90.75/NA & 48.61/78.24 & 42.71/84.61 & 18.60/40.70 & 3.97/14.27  & 38.67/78.74      & 80.06/96.78 & 32.63/63.58  & 47.45/73.92   & 21.57/61.38 & 96.32/99.86 & 80.14/94.99 & 15.34/41.91 & 53.10/90.52 & 63.83/97.51 & 48.15/83.44 \\
            & 8       & 63.23/NA    & 33.33/76.30 & 72.69/NA & 33.33/65.28 & 31.69/70.08 & 11.97/31.10 & 3.81/13.26  & 32.52/72.72      & 78.35/96.53 & 25.19/60.24  & 42.07/71.33   & 16.24/49.83 & 96.75/99.86 & 74.68/93.08 & 11.26/34.29 & 44.76/85.94 & 53.18/95.46 & 39.65/75.82 \\
            & 4       & 57.29/NA    & 19.44/68.06 & 69.16/NA & 10.65/40.28 & 21.26/55.87 & 6.22/19.31  & 3.44/12.43  & 28.53/63.19      & 72.16/94.61 & 23.22/53.57  & 34.64/64.44   & 11.57/39.00 & 95.62/99.72 & 68.27/90.67 & 8.50/26.74  & 34.51/75.27 & 43.15/94.61 & 28.32/73.64 \\
            & 2       & 39.01/NA    & 25.00/65.28 & 41.85/NA & 3.70/17.59  & 14.51/42.98 & 3.01/11.32  & 2.75/10.01  & 23.17/47.03      & 62.79/88.02 & 13.81/38.09  & 27.02/55.71   & 8.24/29.33  & 88.12/99.58 & 58.38/85.34 & 4.75/21.11  & 25.21/65.79 & 36.07/86.58 & 20.26/55.77 \\\hline
\multirow{4}*{\textbf{Swin}}   & 16      & 48.99/NA    & 26.39/72.92 & 70.48/NA & 50.00/77.78 & 33.36/73.46 & 18.16/41.22 & 3.68/13.00  & 44.80/84.01      & 81.42/96.48 & 23.60/67.98  & 43.43/71.21   & 19.43/54.08 & 86.56/99.01 & 66.12/88.26 & 18.53/45.91 & 53.99/93.03 & 41.28/89.82 & 50.11/89.11 \\
            & 8       & 44.73/NA    & 22.22/62.50 & 60.35/NA & 30.09/61.11 & 21.74/59.32 & 0.96/4.42   & 3.45/11.80  & 36.57/75.08      & 76.79/95.57 & 27.01/60.55  & 37.84/65.73   & 11.98/39.78 & 84.02/99.43 & 57.49/82.30 & 9.42/32.73  & 6.91/45.65  & 25.37/87.28 & 38.13/82.14 \\
            & 4       & 46.52/NA    & 16.67/60.42 & 75.33/NA & 15.28/45.83 & 12.77/42.69 & 5.98/19.07  & 2.70/9.88   & 25.05/62.56      & 72.96/94.41 & 17.75/46.74  & 31.29/58.99   & 8.29/29.54  & 70.58/97.88 & 44.67/71.26 & 8.47/25.22  & 23.30/62.63 & 22.37/84.48 & 24.62/65.58 \\
            & 2       & 34.87/NA    & 20.14/58.33 & 62.56/NA & 4.63/18.52  & 11.81/34.77 & 3.41/13.13  & 2.06/8.14   & 22.38/51.02      & 61.83/88.52 & 13.20/39.00  & 23.49/49.86   & 6.59/24.81  & 74.68/98.30 & 38.58/64.78 & 6.52/22.42  & 20.56/51.19 & 19.09/84.99 & 17.43/54.68\\\hline
\end{tabular}}
\end{sidewaystable}

\begin{sidewaystable}[]
\caption{Top-1 and top-5 accuracy of few-shot learning based on self-supervised pre-training on training split3. NL means NonLocal network and TSF means TimeSformer.}\label{tab:ssl res3}
\centering
\scalebox{0.5}{
\begin{tabular}{c|c|cccccccccccccccccc}
     & \# Shot & XD-Violence & UCF-Crime   & MUVIM    & WLASL       & Jester      & UAV Human   & CharadesEGO & Toyota Smarthome & Mini HACS   & MPII Cooking & Mini Sports1M & FineGym     & MOD20       & COIN        & MECCANO     & InHARD      & PETRAW      & MISAW       \\\hline
\multirow{4}*{\textbf{TSN}}  & 16      & 47.53/NA    & 31.25/73.61 & 58.59/NA & 2.78/7.87   & 8.72/30.80  & 1.69/7.68   & 2.99/10.98  & 14.47/45.55      & 56.55/86.05 & 18.21/42.34  & 33.00/63.43   & 9.25/36.02  & 83.73/99.15 & 54.76/83.31 & 9.03/31.95  & 9.71/41.78  & 25.73/69.29 & 25.93/71.24 \\
     & 8       & 33.30/NA    & 20.14/59.72 & 55.07/NA & 1.39/5.09   & 4.58/21.26  & 0.73/3.54   & 2.16/8.45   & 10.90/33.39      & 49.85/81.92 & 12.44/38.09  & 25.11/55.81   & 5.43/24.71  & 79.49/98.87 & 39.72/73.54 & 6.31/23.91  & 7.45/52.15  & 4.21/65.73  & 8.93/46.62  \\
     & 4       & 43.72/NA    & 16.67/57.64 & 47.14/NA & 1.39/2.78   & 4.43/19.94  & 0.87/3.43   & 1.70/6.03   & 13.84/33.26      & 39.93/73.01 & 4.25/27.47   & 16.76/43.04   & 5.13/21.89  & 68.60/99.01 & 27.22/60.47 & 1.20/18.63  & 25.21/43.50 & 2.05/70.83  & 10.24/28.98 \\
     & 2       & 43.72/NA    & 12.50/55.56 & 65.64/NA & 0.46/3.70   & 4.41/19.95  & 0.73/3.43   & 1.00/4.56   & 7.95/20.12       & 26.33/59.87 & 6.22/20.03   & 9.45/28.81    & 3.86/13.57  & 63.08/94.48 & 13.52/38.64 & 9.78/24.76  & 5.90/26.58  & 17.09/61.11 & 3.70/43.14  \\\hline
\multirow{4}*{\textbf{TSM}}  & 16      & 56.61/NA    & 33.33/79.17 & 82.82/NA & 20.83/49.07 & 24.37/63.27 & 8.99/27.43  & 2.98/11.30  & 35.73/84.63      & 54.73/85.60 & 32.02/69.65  & 34.02/63.43   & 16.32/52.97 & 85.71/99.15 & 57.74/84.71 & 13.18/41.02 & 33.31/83.19 & 52.08/94.10 & 47.93/91.50 \\
     & 8       & 51.35/NA    & 26.39/79.17 & 84.14/NA & 8.33/29.17  & 10.51/33.29 & 2.70/10/05  & 2.33/9.23   & 26.49/64.79      & 48.49/81.82 & 18.97/58.73  & 27.19/55.46   & 9.33/37.84  & 83.03/98.02 & 46.19/77.54 & 10.77/32.91 & 15.49/59.00 & 34.25/88.97 & 31.59/73.86 \\
     & 4       & 41.14/NA    & 19.44/68.06 & 67.84/NA & 2.31/5.56   & 5.12/21.78  & 0.88/3.99   & 2.25/8.00   & 11.19/38.78      & 40.94/74.47 & 11.23/38.85  & 19.20/45.81   & 6.10/25.59  & 75.67/97.60 & 35.34/67.20 & 7.58/27.95  & 7.39/26.82  & 24.63/80.37 & 14.38/52.51 \\
     & 2       & 36.32/NA    & 15.28/63.19 & 53.74/NA & 1.39/4.63   & 4.79/21.51  & 0.88/3.58   & 1.24/5.48   & 5.61/35.52       & 32.43/65.86 & 5.61/15.17   & 13.39/35.17   & 4.84/20.10  & 68.32/97.31 & 23.73/54.44 & 8.71/26.96  & 6.73/37.25  & 27.50/78.55 & 19.61/59.91 \\\hline
\multirow{4}*{\textbf{I3D}}  & 16      & 54.82/NA    & 29.86/77.78 & 87.22/NA & 27.31/62.50 & 29.67/72.63 & 6.10/20.66  & 3.17/11.89  & 37.35/81.52      & 62.84/89.84 & 19.88/54.17  & 36.69/65.11   & 14.08/45.46 & 88.13/99.58 & 65.42/89.15 & 15.80/40.84 & 45.41/88.14 & 46.23/92.10 & 48.37/91.94 \\
     & 8       & 54.15/NA    & 31.94/69.06 & 74.01/NA & 13.89/37.96 & 12.45/39.93 & 1.60/6.40   & 2.50/9.52   & 28.79/69.46      & 57.65/86.05 & 20.79/56.60  & 29.88/58.69   & 7.81/31.00  & 82.04/99.86 & 54.31/84.33 & 9.28/30.82  & 25.74/78.43 & 24.42/85.99 & 35.51/83.01 \\
     & 4       & 52.13/NA    & 21.53/70.83 & 65.2/NA  & 2.78/7.41   & 6.19/24.24  & 1.03/3.78   & 2.10/7.90   & 19.38/45.43      & 45.17/77.84 & 17.60/44.92  & 21.66/48.48   & 6.06/23.67  & 74.82/98.16 & 42.07/72.53 & 5.21/21.75  & 13.23/44.28 & 23.01/83.63 & 21.35/57.08 \\
     & 2       & 40.47/NA    & 16.67/65.28 & 33.48/NA & 1.85/5.56   & 4.56/21.43  & 0.73/3.71   & 1.41/5.68   & 9.87/37.90       & 27.19/59.47 & 8.50/31.11   & 13.12/32.94   & 5.29/20.59  & 63.79/95.47 & 25.38/59.45 & 5.14/22.92  & 2.92/27.41  & 25.94/83.35 & 14.81/39.43 \\\hline
\multirow{4}*{\textbf{NL}}   & 16      & 55.04/NA    & 30.56/81.94 & 92.07/NA & 32.87/65.28 & 32.43/74.07 & 5.80/19.50  & 3.32/12.30  & 32.47/78.63      & 61.83/88.97 & 29.29/65.40  & 36.20/64.66   & 12.36/42.79 & 86.14/99.58 & 66.12/88.83 & 14.53/38.67 & 39.39/84.62 & 40.69/93.38 & 48.37/88.67 \\
     & 8       & 48.54/NA    & 25.69/72.92 & 81.5/NA  & 9.26/35.19  & 14.17/41.59 & 1.00/4.75   & 2.44/9.39   & 23.39/60.83      & 54.78/84.74 & 20.94/58.27  & 28.46/57.58   & 7.71/31.38  & 78.22/98.30 & 54.31/83.38 & 11.26/32.66 & 10.67/61.62 & 28.99/84.40 & 30.72/72.98 \\
     & 4       & 47.98/NA    & 21.53/68.75 & 84.15/NA & 3.70/6.48   & 6.57/25.46  & 0.89/3.62   & 1.97/7.52   & 18.04/43.05      & 44.46/77.49 & 16.54/40.06  & 21.05/47.47   & 5.84/22.68  & 69.87/96.89 & 39.09/72.21 & 5.92/22.49  & 9.89/38.92  & 20.73/84.09 & 15.90/52.29 \\
     & 2       & 41.48/NA    & 18.06/60.42 & 48.02/NA & 1.85/5.56   & 5.96/22.64  & 0.68/3.55   & 1.14/5.37   & 7.32/31.11       & 27.54/56.50 & 4.25/25.95   & 13.49/33.47   & 4.83/20.65  & 63.51/93.21 & 23.53/55.96 & 5.63/24.02  & 1.43/24.85  & 18.91/70.73 & 13.94/44.88 \\\hline
\multirow{4}*{\textbf{TSF}}   & 16      & 58.97/NA    & 32.64/76.39 & 66.52/NA & 6.94/23.15  & 14.45/40.81 & 4.46/14.73  & 3.42/13.15  & 25.93/61.48      & 67.62/92.90 & 22.00/49.77  & 35.95/65.79   & 10.63/39.24 & 86.56/99.01 & 58.50/86.55 & 9.42/28.91  & 21.93/67.82 & 43.28/89.76 & 25.49/75.82 \\
     & 8       & 43.95/NA    & 20.83/72.92 & 54.19/NA & 3.70/12.50  & 7.12/27.60  & 1.50/6.18   & 2.68/10.67  & 20.60/48.76      & 61.68/90.58 & 11.23/34.75  & 29.54/59.65   & 7.37/28.75  & 85.01/99.01 & 47.70/79.95 & 8.82/26.14  & 13.17/52.62 & 37.53/87.87 & 14.38/54.47 \\
     & 4       & 44.17/NA    & 18.75/62.50 & 66.52/NA & 1.39/3.70   & 5.62/22.72  & 1.01/4.31   & 2.37/8.78   & 17.21/42.08      & 52.47/83.74 & 6.37/29.89   & 22.59/50.97   & 5.92/24.86  & 78.50/97.88 & 36.93/69.92 & 4.68/22.17  & 14.00/43.98 & 32.14/91.79 & 12.85/53.81 \\
     & 2       & 38.34/NA    & 18.06/56.25 & 33.48/NA & 1.39/4.63   & 4.77/21.58  & 0.79/3.62   & 1.61/6.45   & 11.93/37.55      & 39.48/69.34 & 5.16/21.70   & 15.32/38.93   & 4.48/17.06  & 67.47/89.96 & 27.79/58.63 & 3.86/17.50  & 8.64/51.43  & 21.34/81.97 & 16.12/52.94 \\\hline
\multirow{4}*{\textbf{Swin}} & 16      & 56.61/NA    & 34.03/77.78 & 78.85/NA & 21.76/49.07 & 24.70/59.58 & 12.97/33.92 & 3.43/11.64  & 33.37/79.62      & 69.49/92.80 & 30.35/62.97  & 38.13/66.14   & 15.41/47.86 & 88.97/99.58 & 60.98/86.04 & 17.07/45.77 & 34.56/77.95 & 31.99/83.27 & 34.64/88.02 \\
     & 8       & 49.33/NA    & 23.61/69.44 & 70.93/NA & 10.19/39.81 & 13.49/40.58 & 6.08/19.07  & 2.85/9.45   & 26.62/65.14      & 60.37/88.62 & 27.16/53.41  & 30.70/58.44   & 9.65/35.73  & 83.88/98.59 & 47.91/77.09 & 8.32/28.80  & 23.30/66.57 & 18.86/80.30 & 22.66/68.41 \\
     & 4       & 55.04/NA    & 17.36/65.97 & 74.45/NA & 2.78/11.57  & 6.24/25.56  & 2.93/10.70  & 2.13/7.79   & 23.58/52.57      & 49.09/78.90 & 18.36/42.49  & 22.40/47.68   & 7.08/26.83  & 75.81/97.74 & 35.79/63.07 & 5.46/21.71  & 18.71/55.60 & 26.04/82.32 & 18.95/60.35 \\
     & 2       & 41.93/NA    & 20.14/56.94 & 48.02/NA & 1.39/7.41   & 5.02/21.46  & 1.33/5.54   & 1.48/5.61   & 17.19/43.46      & 37.66/66.06 & 13.20/35.81  & 16.06/37.82   & 5.40/20.40  & 69.87/95.47 & 26.14/52.86 & 4.53/19.84  & 10.79/40.17 & 19.93/74.86 & 11.76/47.28\\\hline
\end{tabular}}
\end{sidewaystable}

\clearpage
\section{Unsupervised domain adaptation}
\label{appendix uda}
In BEAR, we construct two different types of transfer for unsupervised domain adaptation (UDA): inter-dataset transfer and intra-dataset transfer. We construct paired source-target with different datasets for inter-dataset, while we build paired data within one dataset according to similar or same actions for intra-dataset. The main challenge in our UDA benchmark could be caused by viewpoint change (e.g., ToyotaSmarthome-MPII-Cooking), long-tail problem (PHAV-Mini-Sports1M), etc.

\subsection{Inter-dataset}
\begin{figure}[htbp]
\centering
\subcaptionbox{\label{1}}{\includegraphics[width = .48\linewidth]{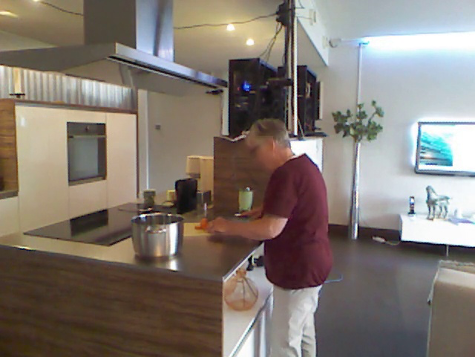}}
\subcaptionbox{\label{2}}{\includegraphics[width = .48\linewidth]{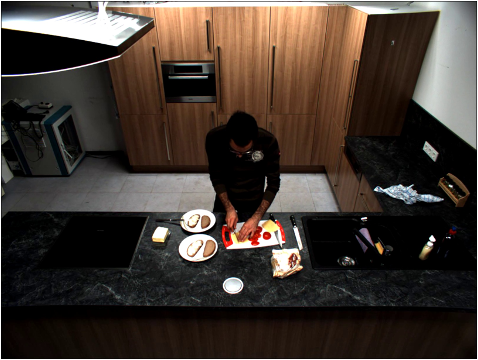}}
\caption{Example frames of Toyota Smarthome and MPII-Cooking. The left frame is from Toyota Smarthome, in which the videos are captured from 7 different cameras; the right one is from MPII-Cooking and is recorded by a fixed down view camera. }
\label{example_figs} 
\end{figure}

\begin{figure*}[htbp]
\centering
    \includegraphics[width=0.5\linewidth]{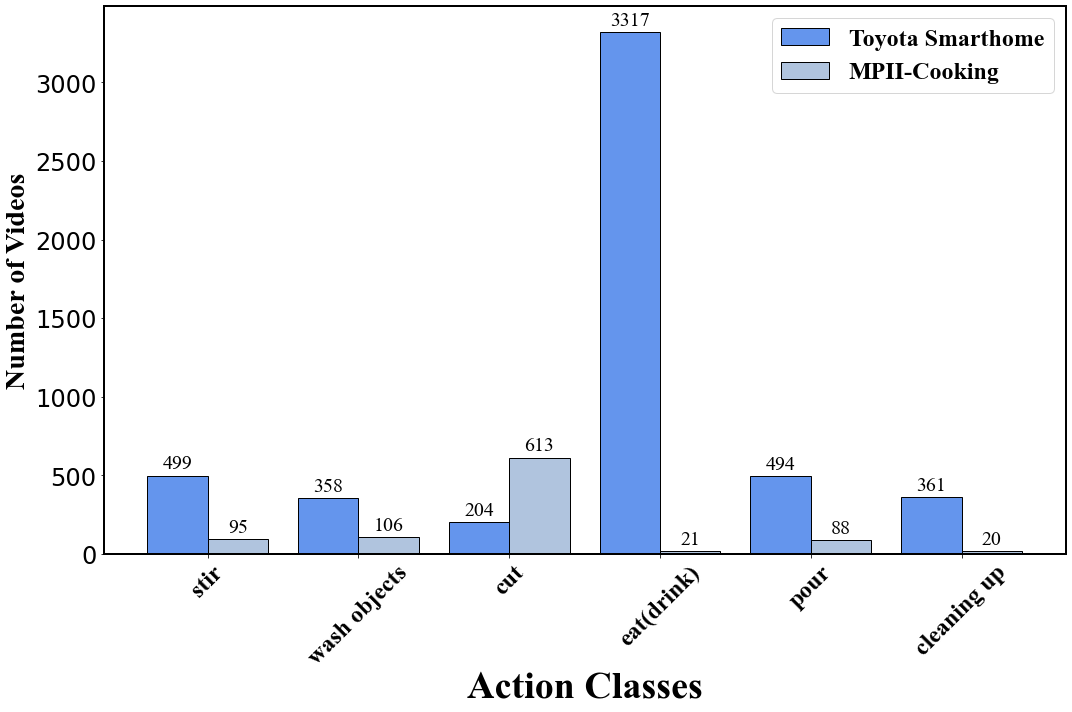}
    \caption{Class-wise distribution of videos in Toyota Smarthome-MPII-Cooking. There is a  severe long-tail distribution  in Toyota Smarthome.}
    \label{daily}
\end{figure*}

\paragraph{ToyotaSmarthome-MPIICooking}
One of the features of our benchmark is that we collect several datasets with obvious viewpoint shifts, and we also leverage this when we build our UDA datasets. As shown in Figure~\ref{example_figs},  Toyota Smarthome and MPII-Cooking consists of videos from different viewpoints. Specifically, as shown in Table~\ref{toyota-mpii}, we select 6 common categories in Toyota Smarthome and MPII-Cooking to construct the new Toyota Smarthome-MPII-Cooking dataset.
As shown in Figure~\ref{daily}, the video numbers can be imbalanced across source data and target one, for action class \textit{'eat(drink)'}, there are a total of 3317 videos in Toyota Smarthome, since 7 original classes are merged; while there are only 21 samples from the original class \textit{'taste'} in MPII-Cooking. The number of  videos is 5,233 and 943 for Toyota Smarthome and MPII-Cooking, respectively. 

\begin{figure*}[htbp]
\centering
    \includegraphics[width=0.8\linewidth]{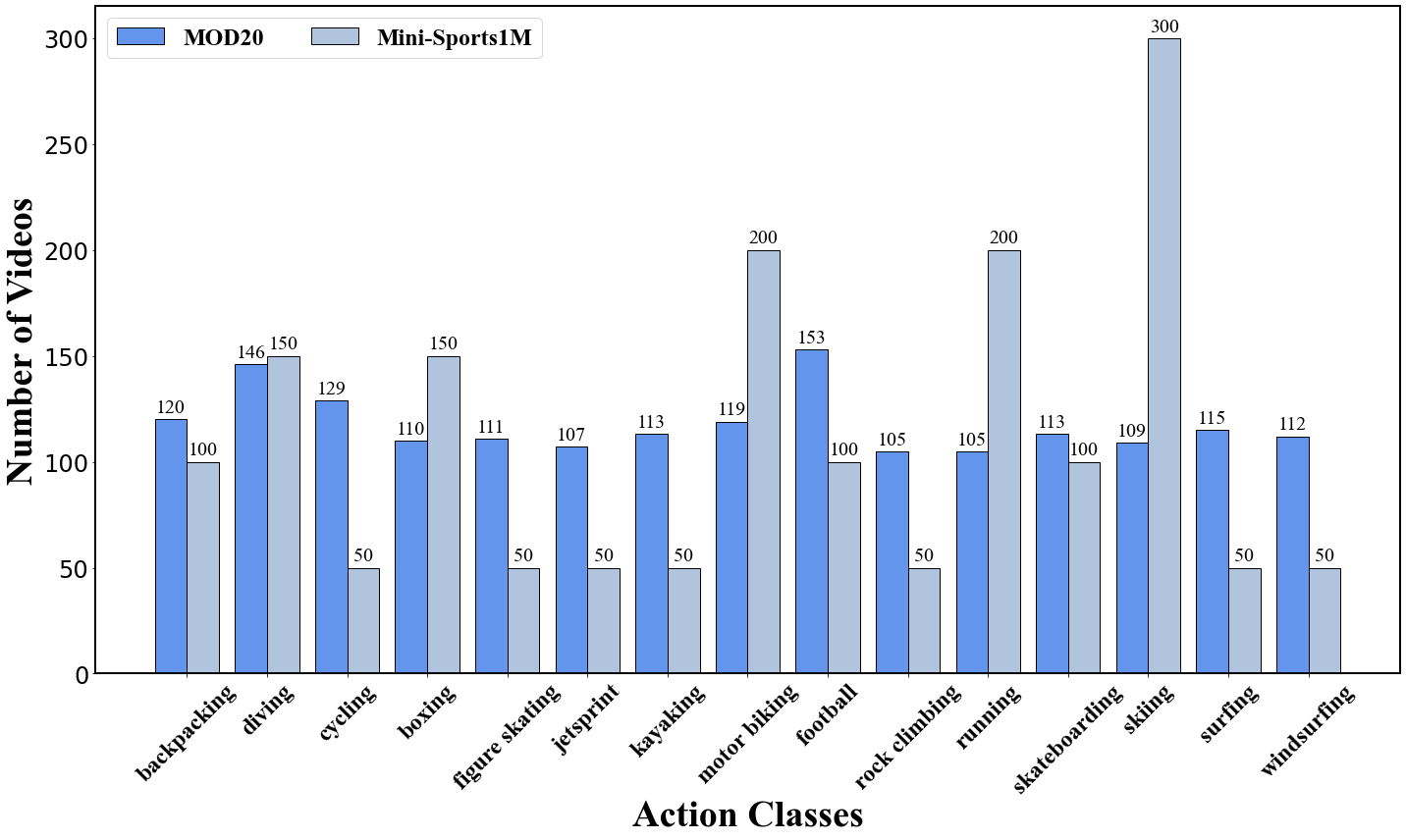}
    \caption{Class-wise distribution of videos in Mini-Sports1M-MOD20. Video numbers in this dataset are much more balanced for source and target.}
    \label{sports}
\end{figure*}

\vspace{-0.5 em}
\paragraph{Mini-Sports1M-MOD20}
Similarly, as shown in Table~\ref{sports-mod}, for Mini-Sports1M and MOD20, we select 15 categories to build the UDA dataset. The statistic is shown in  Figure~\ref{sports}. 
In contrast to Toyota Smarthome-MPII-Cooking, the data distribution in Mini-Sports1M-MOD20 is much more balanced. There are 1,650 videos for Mini-Sports1M and 1,767 for MOD20.

\begin{figure*}[htbp]
\centering
    \includegraphics[width=0.5\linewidth]{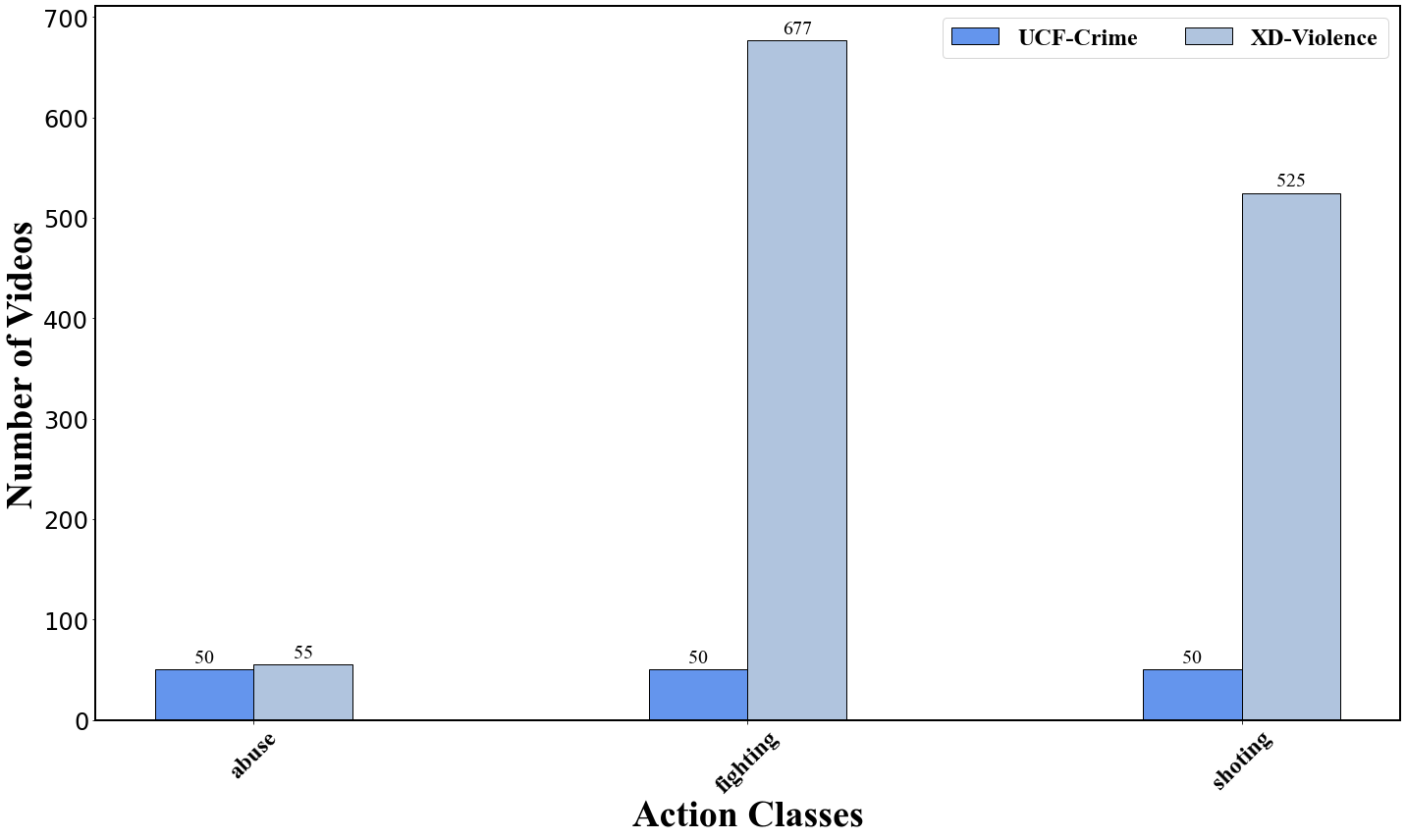}
    \caption{Class-wise distribution of videos in UCF-Crime-XD-Violence.}
    \vspace{3 em}
    \label{anomaly}
\end{figure*}

\paragraph{UCF-Crime-XD-Violence}
Similarly, UCF-Crime and XD-Violence share three classes: \textit{abuse}, \textit{fighting}, and \textit{shooting}. As shown in Figure~\ref{anomaly}, sample numbers of \textit{fighting} and \textit{shooting} showcase an obvious imbalance distribution, which makes the UDA task here much more challenging.

\begin{figure*}[htbp]
\centering
    \includegraphics[width=0.5\linewidth]{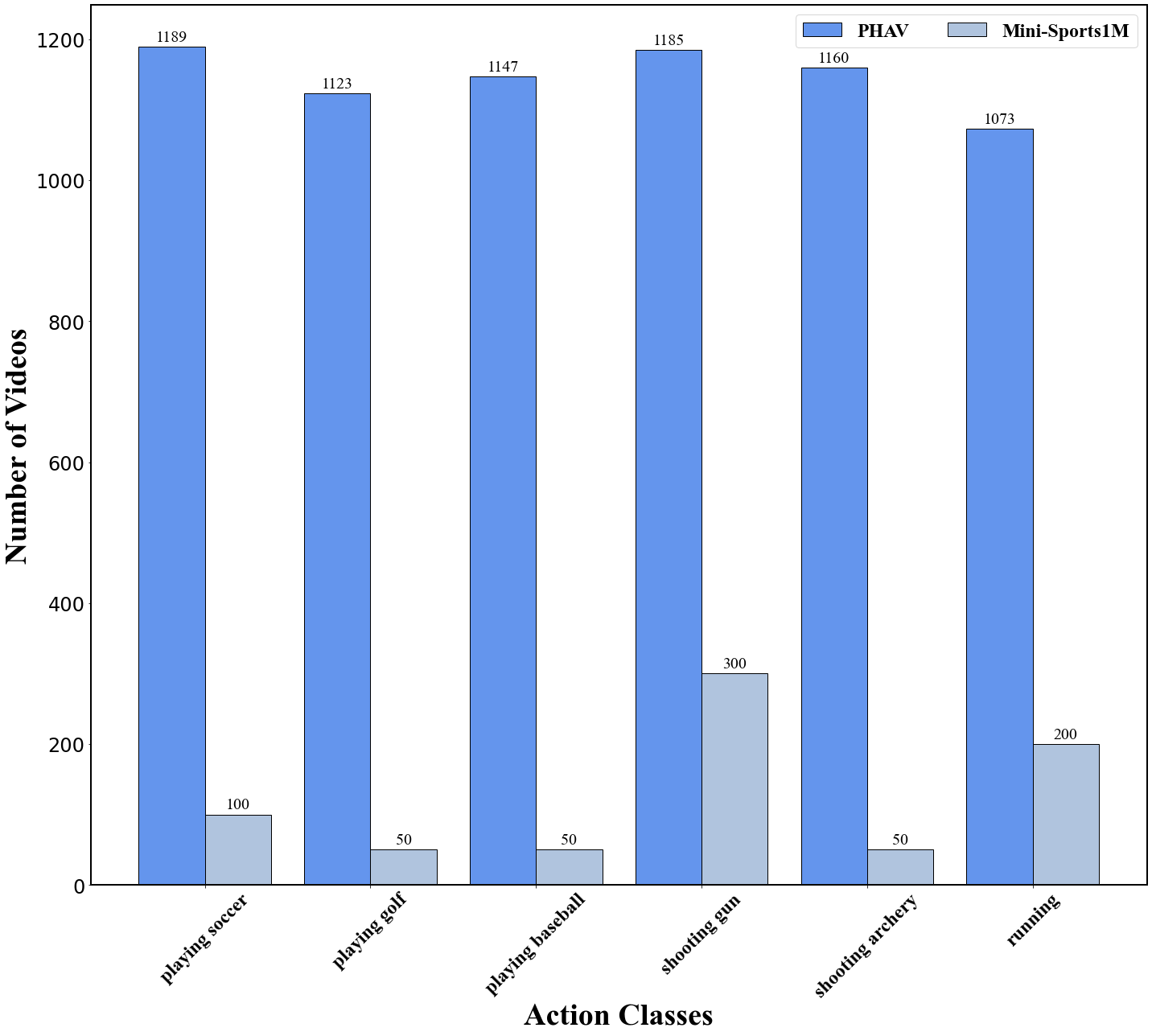}
    \caption{Class-wise distribution of videos in PHAV-Mini-Sports1M.}
    \vspace{6 em}
    \label{phav-ms}
\end{figure*}

\paragraph{PHAV-Mini-Sports1M}
We also consider the synthetic-to-real transfer and we leverage an existing dataset PHAV~\cite{roberto2017procedural}. As shown in Table~\ref{PM}, we combine 15 classes from Mini-Sports1M into 6 categories (\textit{playing soccer}, \textit{playing golf}, \textit{playing baseball}, \textit{shooting gun}, \textit{shooting archery} and \textit{running}) existing in PHAV to build the paired dataset. We also illustrate the class-wise distribution of this dataset in Figure~\ref{phav-ms}. PHAV contains much more samples than Mini-Sports1M due to it is easily generated.

\subsection{Intra-dataset}
\begin{figure*}[t]
\centering
    \includegraphics[width=0.5\linewidth]{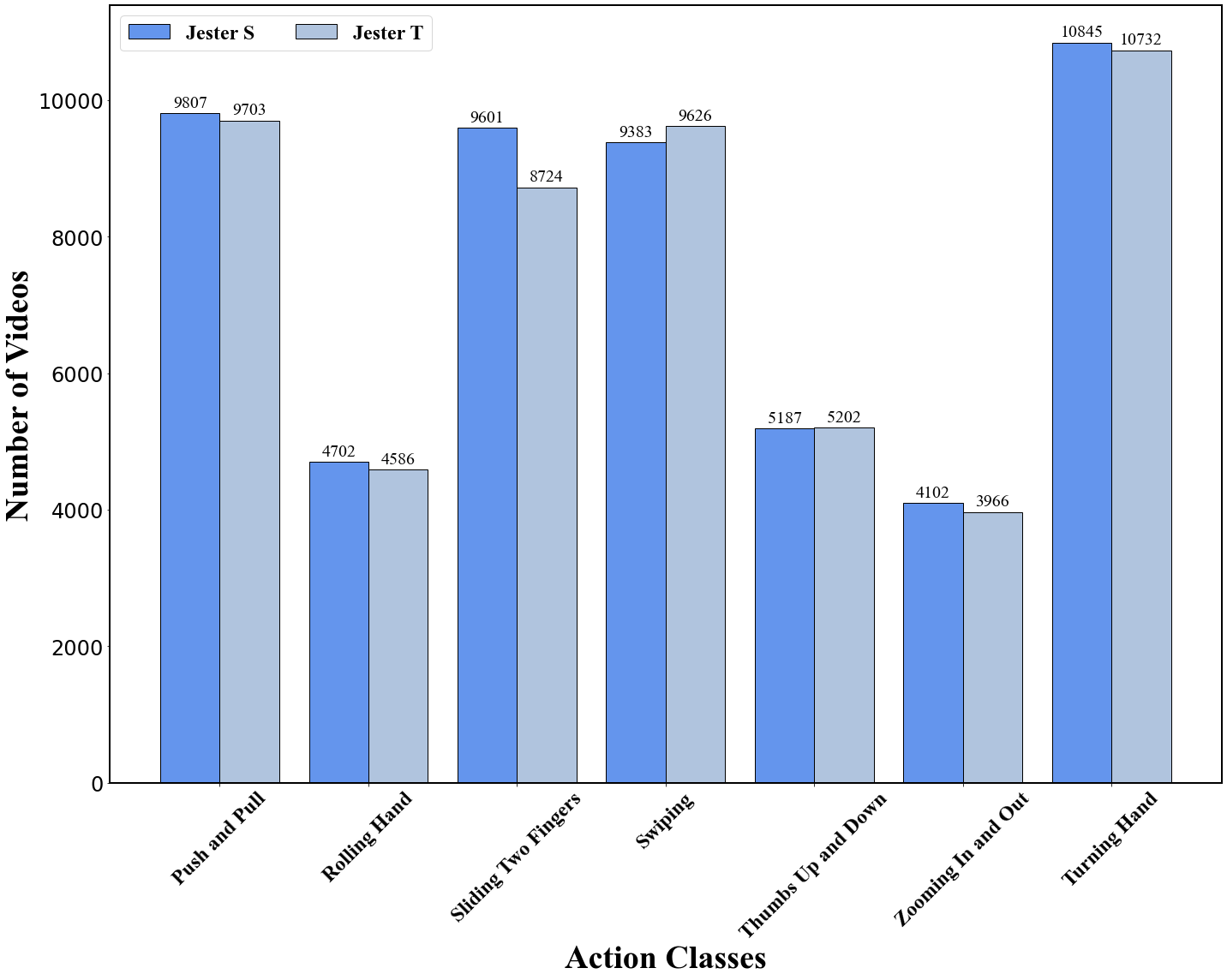}
    \caption{Class-wise distribution of videos in Jester(S-T).}
    \vspace{-3 em}
    \label{jester-st}
\end{figure*}

\paragraph{Jester(S-T)} 
We also include existing Jester(S-T)\cite{sahoo2021contrast} in BEAR. The category information is shown in Table~\ref{jester} that each identical action with a contrary direction is merged into one category. For completeness, we also include its class-wise distribution in Figure~\ref{jester-st}.

\begin{figure}[htbp]
\centering
\subcaptionbox{\label{1}}{\includegraphics[width = .32\linewidth]{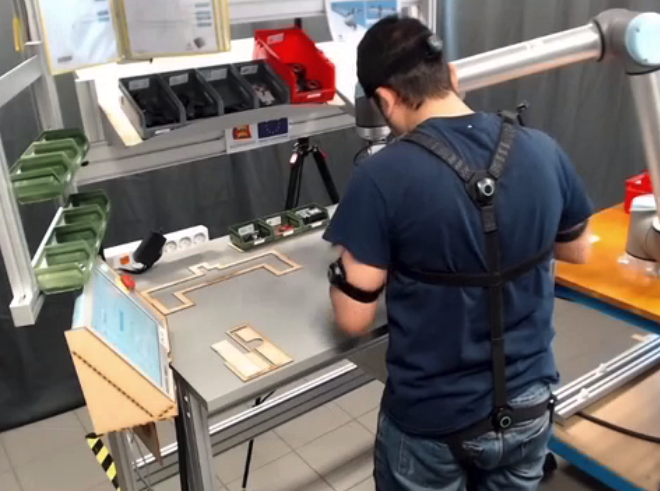}}
\subcaptionbox{\label{2}}{\includegraphics[width = .32\linewidth]{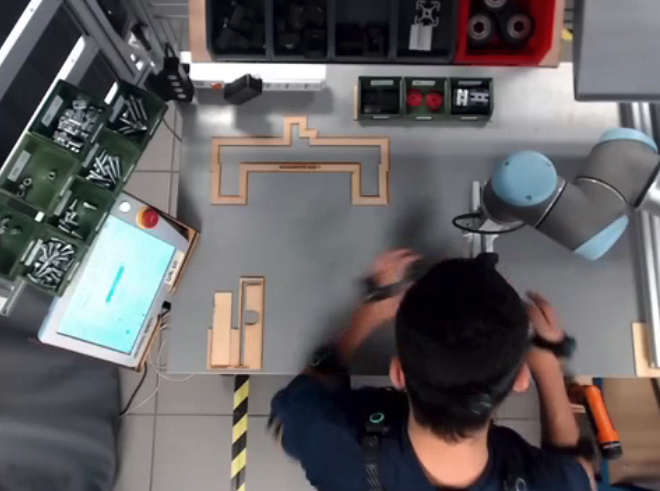}}
\subcaptionbox{\label{2}}{\includegraphics[width = .32\linewidth]{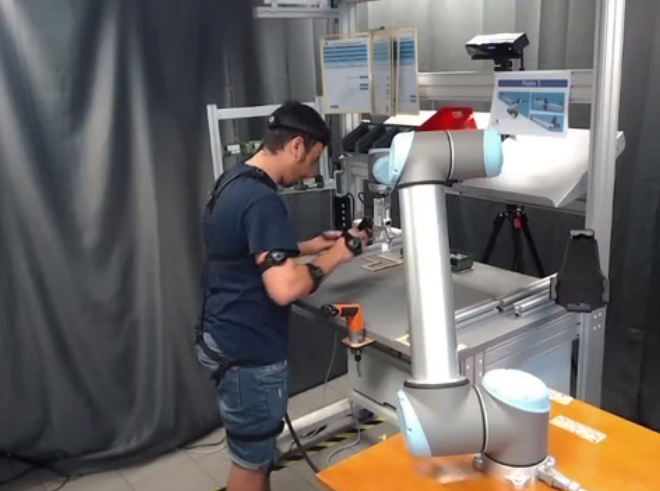}}
\caption{Examples of InHARD(Left-Top-Right). The shown frames are from the left, top, and right views, respectively. The left view can be severely occluded, making it much more challenging to transfer knowledge from this view to others.}
\label{inhard-view} 
\end{figure}

\begin{figure*}[htbp]
\centering
    \includegraphics[width=0.8\linewidth]{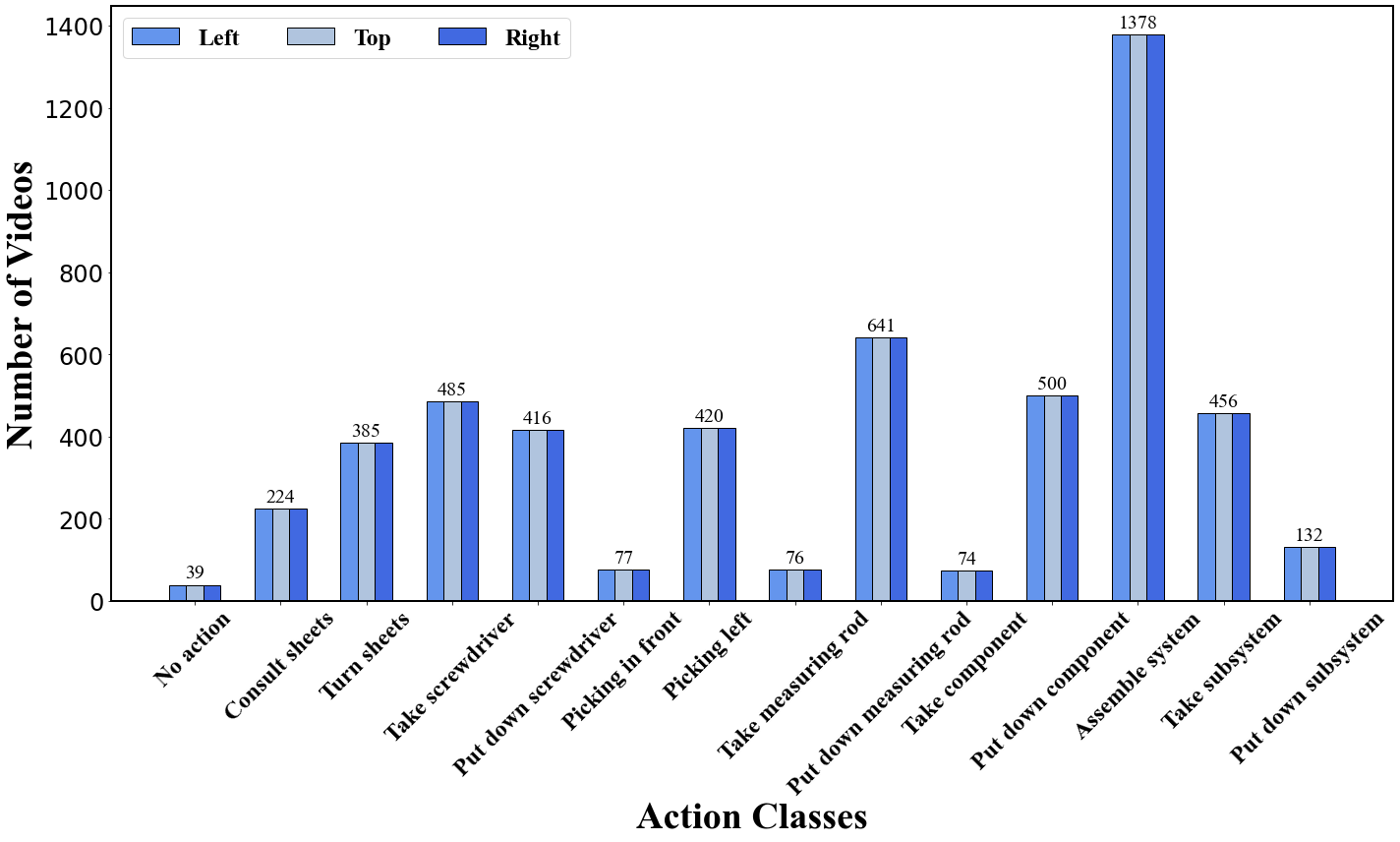}
    \caption{Class-wise distribution of videos in InHARD(Left-Top-Right).}
    \label{inhard_da}
\end{figure*}

\paragraph{InHARD(Left-Top-Right)}
InHARD naturally contains three different views and each frame contains the top, left, and right views, respectively as shown in Figure~\ref{inhard-view}. We simply split the frames and the category is the same as the original dataset, and the samples in each category are also the same in Figure~\ref{inhard_da}.

\begin{table*}[h]
\centering
\caption{Action classes in Toyota Smarthome-MPII-Cooking.}
\scalebox{0.8}{
\setlength{\tabcolsep}{1mm}{
\begin{tabu}{c|[1.5pt]c|c}
\tabucline[1.5pt]{}
Toyota Smarthome-MPII-Cooking & Toyota Smarthome & MPII-Cooking \\ \tabucline[1.5pt]{}
stir & Cook.Stir & stir \\ \hline
wash objects & Cook.Cleandishes & wash objects \\ \hline
cut & \begin{tabular}[c]{@{}c@{}}Cook.Cut\\ Cutbread\end{tabular} & \begin{tabular}[c]{@{}c@{}}cut out inside\\cut apart\\cut in\\cut dice\\cut slices\\cut off ends\\cut stripes\end{tabular} \\ \hline
eat(drink) & \begin{tabular}[c]{@{}c@{}}Eat.Snack\\Eat.Attable\\Drink.Fromcan\\Drink.Frombottle\\Drink.Fromcup\\cut off ends\\Drink.Fromglass\end{tabular} & taste \\ \hline
pour &\begin{tabular}[c]{@{}c@{}}Pour.Fromkettle\\Pour.Fromcan\\Pour.Frombottle\\Makecoffee.Pourwater\\Makecoffee.Pourgrains\end{tabular} & pour \\\hline
cleaning up & Cook.Cleanup & wipe clean   \\\tabucline[1.5pt]{}
\end{tabu}}}
\label{toyota-mpii}
\end{table*}

\begin{table*}[h]
\centering
\caption{Action classes in Mini-Sports1M-MOD20.}
\scalebox{0.8}{
\setlength{\tabcolsep}{1mm}{
\begin{tabu}{c|[1.5pt]c|c}
\tabucline[1.5pt]{}
Mini-Sports1M-MOD20  & Mini-Sports1M & MOD20 \\ \tabucline[1.5pt]{}
backpacking &  \begin{tabular}[c]{@{}c@{}}backpacking(wilderness)\\ hiking\end{tabular} & backpacking \\ \hline
diving &  \begin{tabular}[c]{@{}c@{}}diving\\free-diving\\cuba diving\end{tabular} & clif$\_$jumping \\ \hline
cycling &  cycling & cycling \\ \hline
boxing &  \begin{tabular}[c]{@{}c@{}}boxing\\shoot boxing\\kick boxing\end{tabular} & motorbiking \\ \hline
figure skating &  figure skating & figure skating \\ \hline
jetsprint &  jetsprint & jetskii \\ \hline
kayaking &  kayaking & kayaking \\ \hline
motor biking & \begin{tabular}[c]{@{}c@{}}motorcycle racing\\grand prix motorcycle racing\\motorcycle speedway\\motorcycle drag racing\\\end{tabular} & motorbiking \\\hline
football &  \begin{tabular}[c]{@{}c@{}}American football\\ Canadian football\end{tabular} & nfl$\_$catches   \\\hline
rock climbing &  rock$\_$climbing & rock$\_$climbing \\ \hline
running & \begin{tabular}[c]{@{}c@{}}free running\\running\\sprint (running)\\cross country running\\\end{tabular} & running \\\hline
skateboarding & \begin{tabular}[c]{@{}c@{}}freeboard (skateboard)\\skateboarding\end{tabular} & skateboarding \\\hline
skiing & \begin{tabular}[c]{@{}c@{}}skiing\\alpine skiing\\cross-country skiing\\freestyle skiing\\nordic skiing\\telemark skiing\\\end{tabular} & skiing \\\hline
surfing &  surfing & surfing \\ \hline
windsurfing &  windsurfing & windsurfing \\ \tabucline[1.5pt]{}
\end{tabu}}}
\label{sports-mod}
\end{table*}


\begin{table*}[h]
\centering
\caption{Action classes in PHAV-Mini-Sports1M.}
\scalebox{0.8}{
\setlength{\tabcolsep}{1mm}{
\begin{tabu}{c|[1.5pt]c|c}
\tabucline[1.5pt]{}
PHAV-Mini-Sports1M & PHAV & Mini-Sports1M \\ \tabucline[1.5pt]{}
playing soccer & kick ball  & \begin{tabular}[c]{@{}c@{}}indoor soccer\\ beach soccer\end{tabular} \\ \hline
playing golf &  golf & golf \\ \hline
playing baseball &  swing baseball & baseball \\ \hline
shooting gun & shoot gun & \begin{tabular}[c]{@{}c@{}}shooting sports\\practical shooting\\cowboy action shooting\\clay pigeon shooting\\skeet shooting\\trap shooting\end{tabular}  \\ \hline
shooting archery &  shoot bow & archery \\ \hline
running & run  &  \begin{tabular}[c]{@{}c@{}}free running\\running\\sprint (running)\\cross country running\\\end{tabular} \\\tabucline[1.5pt]{}
\end{tabu}}}
\label{PM}
\end{table*}

\begin{table*}[h]
\centering
\caption{Action classes in Jester(S-T).}
\scalebox{0.8}{
\setlength{\tabcolsep}{1mm}{
\begin{tabu}{c|[1.5pt]c|c}
\tabucline[1.5pt]{}
Jester & Jester Source & Jester Target \\ \tabucline[1.5pt]{}
Push and Pull & \begin{tabular}[c]{@{}c@{}}Pushing Hand Away\\ Pushing Two Fingers Away\end{tabular} & \begin{tabular}[c]{@{}c@{}}Pulling Hand Away\\ Pulling Two Fingers Away\end{tabular} \\ \hline
Rolling Hand & Rolling Hand Forward & Rolling Hand Backward \\ \hline
Sliding Two Fingers & \begin{tabular}[c]{@{}c@{}}Sliding Two Fingers Left\\ Sliding Two Fingers Up\end{tabular} & \begin{tabular}[c]{@{}c@{}}Sliding Two Fingers Right\\ Sliding Two Fingers Down\end{tabular} \\ \hline
Swiping & \begin{tabular}[c]{@{}c@{}}Swiping Left\\ Swiping Up\end{tabular} & \begin{tabular}[c]{@{}c@{}}Swiping Right\\ Swiping Down\end{tabular} \\ \hline
Thumbs Up and Down & Thumbs Up & Thumbs Down   \\\hline
Zooming In and Out & \begin{tabular}[c]{@{}c@{}}Zooming Out with Full Hand\\ Zooming Out with Two Fingers \end{tabular} & \begin{tabular}[c]{@{}c@{}}Zooming In with Full Hand\\ Zooming In with Two Fingers\end{tabular} \\ \hline
Turning Hand & Turning Hand Counterclockwise & Turning Hand Clockwise  \\\tabucline[1.5pt]{}
\end{tabu}}}
\label{jester}
\end{table*}
\clearpage

\clearpage

\end{document}